\documentclass[journal,final]{IEEEtran}
\def\BibTeX{{\rm B\kern-.05em{\sc i\kern-.025em b}\kern-.08em
    T\kern-.1667em\lower.7ex\hbox{E}\kern-.125emX}}
\usepackage{amsmath, amssymb,mathtools}
\usepackage{stmaryrd}
\usepackage{amsthm,cite}
\usepackage{amsfonts}
\usepackage{graphicx,subfigure}
\usepackage{epsfig,color,pifont,enumerate}
\usepackage{setspace}
\graphicspath{{img/}}
\usepackage[mathscr]{eucal}

\usepackage{geometry}
\DeclareMathAlphabet{\mathpzc}{OT1}{pzc}{m}{it}

\renewcommand{\emptyset}{\O}
\newcommand{\rmin}{R_{\min}}
\renewcommand{\phi}{\varphi}
\renewcommand{\epsilon}{\varepsilon}
\renewcommand{\kappa}{\varkappa}
\newcommand{\sgn}{\text{\bf sgn}\,}

\newcommand{\br}{\mathbb{R}}
 \newcommand{\bldr}{\boldsymbol{r}}
 \newcommand{\pr}[2]{\mathbf{Pr}_{#2}\left[ #1\right]}
 \newcommand{\rvis}{R_{\text{\rm vis}}}
 
 \newcommand{\rext}{\mathscr{R}_{\text{ext}}}
 \newcommand{\rop}{R_{\text{\rm op}}}
 \newcommand{\sa}[2]{\sphericalangle \!\big[ #1, #2\big]}

\newtheorem{define}{Definition}[section]

\newtheorem{assumption}{Assumption}[section]
\newtheorem{lemma}{Lemma}[section]
\newtheorem{cor}{Corollary}[section]
\newtheorem{prop}{Proposition}[section]
\newcommand{\smo}{\text{\scriptsize $\mathscr{O}$}}
\newcommand{\md}{\mathpzc{d}}
\newcommand{\w}[1]{\widetilde{#1}}
\newcommand{\mzp}{\mathpzc{p}}
\newcommand{\mzn}{\mathpzc{n}}

\newcommand{\so}{\text{\scriptsize $\mathcal{O}$}}
\newcommand{\spr}[2]{\left\langle #1; #2 \right\rangle}

\newcommand{\ov}[1]{\overline{#1}}

\newcommand{\ve}{\varepsilon}

\renewcommand{\theequation}{\arabic{section}.\arabic{equation}}

\newcommand{\dist}[2]{\text{\bf dist}\, \left[ #1; #2 \right]}

\newcommand{\rin}{\bldr_{\text{in}}}
\newcommand{\bldp}{\boldsymbol{p}}
 \newcommand{\bldc}{\boldsymbol{c}}
 \newcommand{\mathb}{\mathscr{B}}

\title{Geometric Facts Underlying Algorithms of Robot Navigation for Tight Circumnavigation of Group Objects through Singular Inter-Object Gaps}
\date{}
\author{Valerii Chernov and Alexey Matveev\thanks{The authors are with the Department of Mathematics and Mechanics, Saint Petersburg
University, Universitetskii 28, Petrodvoretz, St.Petersburg,
Russia (e-mail: almat1712@yahoo.com, st096786@student.spbu.ru ). }}
\begin{document}

\maketitle

\begin{abstract}
An underactuated nonholonomic Dubins-vehicle-like robot with a lower-limited turning radius travels with a constant speed in a plane, which hosts unknown complex objects. The robot has to approach and then circumnavigate all objects, with maintaining a given distance to the currently nearest of them. So the ideal targeted path is the equidistant curve of the entire set of objects. The focus is on the case where this curve cannot be perfectly traced due to excessive contortions and singularities. So the objective shapes into that of automatically finding, approaching and repeatedly tracing an approximation of the equidistant curve that is the best among those trackable by the robot. The paper presents some geometric facts that are in demand in research on reactive tight circumnavigation
of group objects in the delineated situation.
\end{abstract}

\section{Introduction}
Autonomous safe navigation of a robot along the boundary of an extended object is a fundamental task in mobile robotics.
This task is of interest for, e.g., border patrolling and surveillance \cite{HadGer10,HS11,BeBeKaMa15}, structure inspection and processing \cite{ShKurKuDa19,BeIsoMey22}, bottom following by underwater robotic vehicles \cite{Yuh00,YuGuHa20},
lane following by robotic cars \cite{LeeHan08}, sensor-based navigation inside a tunnel-like environment without a physical contact with its surface
\cite{MaMaSa20,ChZhSuLi22}, and for many other missions, where it is reasonable to travel along the boundary, at least for a while, at a given distance from it.
An example is the classical method of bypassing enroute obstacles; see e.g., \cite{CH05,BoOrOl08,MaSaHoWa15,HoMaSa15rob,MutAndFai16}.
\par
Another examples arise in the emerging area
of autonomous sensor-based circumnavigation of pointwise/extended single/multiple target(s) by using robotic platforms. In such missions, it is needed to approach the targets and to orbit around them with a narrow margin so that they become inscribed into the robot's path.
Among examples of such missions, there are military \cite{BeMcLGoAn02} and civilian \cite{KuSawSam01} reconnaissance and surveillance, protection of VIP objects, tracking migratory schools of fish species \cite{TaShLoCl14}, pursuit, entrapment, and monitoring of threatful objects, guarding the perimeter of an area, improving connectivity of multihop communication networks \cite{ZavEgePap11}, objects inspection, improving situational awareness before making a closer contact, retrieval and fusion of data from/about spatially distributed sensors/spots, targets acquisition and localization \cite{SpIsIEEE05,BiFiAnDoPa10}.
\par
Many approaches to the problem of boundary following have been offered. Some of them aim at driving the robot near the boundary and do not much care about the distance to it.
Meanwhile, perfectly maintaining a targeted distance to the boundary is critical in many missions.
Applicable approaches can be broadly classified based on the type and amount of the consumed sensor data and computational resources.
If rich information about the scene is available, the targeted equidistant curve to the boundary can be pre-computed to some extent, after which
a standard path-tracking controller can be put in use \cite{AdMiYaAg14}. However, these methods typically call for a relatively sophisticated sensor equipment and extensive data processing, which may not be suitable for e.g., small-size robots. The focus on real-time implementation with minimization of the consumed resources have motivated interest to reactive controllers for which the current control output is a reflex-like response to the current observation and, maybe, current values of a few inner variables of the controller.
\par
There is a vast literature on reactive sensor-based boundary following with maintaining a given distance margin. However many of the proposed controllers are fed by an input
whose creation requires computationally expensive sensing, feature extraction, and classification.
For example, reactive vision navigation (see, e.g., \cite{BoAlOli08,Guzel13,QiLiLi22} for surveys) employs the capacity of computer vision systems to capture and memorize a whole chunk of the scene and typically builds on extraction of image features and estimation of their motion within a sequence of images, which calls for intensive image processing. Another examples of perceptually and computationally demanding inputs, not necessarily confined to the area of visual navigation,
can be found in e.g. \cite{BeMaTe00,ZZS09,FK05,MTS11,MHS11,ZJK04,ZhFrPaLuLe07}. A variety of computationally cheap reactive
algorithms is also available. They are based on e.g., sliding-modes \cite{MTS11,JiZhBi12,MaChaSa12a,YaYuLi15,YuGuHa20}, PID regulation \cite{SuTiDe18}, and Lyapunov functions \cite{WaWiSu13,WeDoLi17}. Ideas of fuzzy logics and reinforcement learning are rather popular here \cite{JuaHsu09,JuJhChHs18}, though often not being backed by completed rigorous proofs.
\par
A common limitation of the previous research on reactive sensor-based following unknown boundaries is the assumption that the targeted equidistant curve can actually be traced by the mobile robot. However, this may not be the case, especially for underactuated and nonholonomic robots, which are not capable of making very sharp turns. Then such a robot faces the need to reactively build on the fly and then follow a trackable safe path that optimally (sub-optimally) approximates the equidistant curve. This situation may happen due to a variety of reasons, including, e.g., kinks of the boundary. Another reason is that as the distance to the concave boundary increases, so does the curvature of the equidistant curve, which may result even in the loss of regularity and emergence of singular points on the curve  \cite{Arnold91,Arnold04}. One more reason is switching between parts of a group object.
\par
This paper is aimed at filling this gap. It presents some geometric facts that are
in demand, according to the experience
of the authors, in research on reactive tight circumnavigation of group objects in the case where the targeted equidistant curve cannot be perfectly tracked because of singularities and excessive contortions. The proofs of these geometric facts may need rather tedious, though straightforward, more or less, considerations. This
partly discloses the idea behind the current text: To unload research papers on the above main topic from the need to perform these technical developments.
\par
The remainder of the paper is organized as follows.
Section~\ref{sec1} offers the description of the basic problem.
Section~\ref{sec.ass} focuses on the assumptions of our theoretical analysis and presents the main results of the paper.
Their proofs are scattered over Sections~\ref{app.1}--\ref{app.fence}, which also contain geometric facts that are useful in their own rights in the concerned area.
\par
The following conventions and notations are adopted in the paper:
\begin{itemize}
\item
Positive angles are counted counterclockwise;
\item a curve is said to be {\it regular} if it has a parametric representation with everywhere nonzero derivative;
\item
$\spr{\cdot}{\cdot}$, standard inner product;
\item
$\|\cdot\|$, standard Euclidean norm;
\item
$\pr{\bldr}{E}$, set of the {\it projections} of $\bldr \in \br^2$ onto the closed set $E$, i.e., the minimizers of $\|\bldr - \bldp\|, \bldp \in E$;
\item
$\dist{\bldr}{E} := \|\bldr-\pr{\bldr}{E}\|$, distance from $\bldr$ to $E$;
\item
$\mathscr{N}^\eta(E) := \big\{ \bldr: \dist{\bldr}{E} \leq \eta \big\}$, $\eta$-neighborhood of the set $E$;
\item
$\partial E$, boundary of the set $E$;
\item $\mathbf{int}E$, collection of all interior points of the set $E$;
\item  ``$R$-circle '' means ``circle with a radius of $R$'';
\item  ``$R$-disc'' means ``closed disc with a radius of $R$'';
\item
$[\bldp_1,\bldp_2]$, closed straight line segment that bridges the points $\bldp_1$ and $\bldp_2$;
\item
$[\bldp_1,\bldp_2)$, line segment $[\bldp_1,\bldp_2]$ deprived of $\bldp_2$.
\end{itemize}

\section{Problem of tight circumnavigation of a group object}\label{sec1}
A non-holonomic under-actuated planar robot travels with a constant speed $v >0$ and is driven by the
angular velocity $\omega$ limited in absolute value by a constant $\ov{\omega} >0$.
Such robots are classically described by the {\it Dubins-car model}:
\begin{equation}
\label{ch3:1}
\begin{array}{l}
\dot{\bldr} = v \vec{e}(\theta), \; \vec{e}(\theta) := \left[ \begin{smallmatrix} \cos \theta \\ \sin \theta \end{smallmatrix}\right],
\\
\dot{\theta} = \omega \in [-\overline{\omega}, \overline{\omega}],
\end{array} \;
\begin{array}{l}
\bldr(0) = \rin,
\\
\theta(0) = \theta_{\text{in}}.
\end{array}
\end{equation}
Here $\bldr$ and $\theta$ give the robot's location and orientation, respectively.
Equations \eqref{ch3:1} capture the robot's capacity to move over paths
whose curvature radius $\geq \rmin = \frac{v}{\overline{\omega}}$.
\par
The plane hosts $N$ unknown {\it objects}; the $i$th of them occupies a set $D_i$.
In its local frame and within a given {\it visibility distance} $\rvis$, the robot has access to the part of the scene
that is in direct line of sight.
\par
The objective is to find, approach and then repeatedly circumnavigate the union $D:= \bigcup_iD_i$ of the objects. For the best fulfilment of the mission, the robot should go close enough to the objects: it should adhere to a ``preferable'' distance $d_0$ to them as much as possible. (Here $d_0 > \rmin$ to reserve room for a collision avoidance turn, if necessary.)
The best way to comply with this demand is to trace the boundary $\mathpzc{B}:=\partial \mathscr{N}^{d_0}(D)$ of the $d_0$-neighborhood of $D$.
Three impediments may obstruct implementation of this decision:
\begin{enumerate}[{\bf a)}]
\item $\mathpzc{B}$ may contain disconnected pieces;
\item At some places, the curvature radius of $\mathpzc{B}$ may be less than the minimal turning radius $\rmin$ of the robot;
\item Following $\mathpzc{B}$ may result in a situation where the robot is stuck inside a narrow ``cave''.
\end{enumerate}
\par
The possibility of a) motivates the following.
\begin{assumption}
\label{ass.union}
The boundary $\partial \mathscr{N}^{d_0}(D)$ contains a Jordan loop (i.e., a closed non self-intersecting curve) $\mathscr{P}$ such that all objects lie on the same side of it.
\end{assumption}
\par
This loop is called the {\it outer perimeter}. By the Jordan curve theorem, $\mathscr{P}$ splits the plane into two regions; the one $\rext$ that is free of the objects is called the {\it exterior} of the outer perimeter. To cope with the likelihood of a), the goal is preliminarily specified as
arriving at and then tracing the outer perimeter, if starting in its exterior.
\par
The situation from c) occurs
inside a ``cave'' in $\rext$ that is less in width
than the diameter $2 \rmin$ of the sharpest feasible $U$-turn or, equivalently, is such that no $\rmin$-disc can be placed in this cave.
Though a chance to turn back may appear deeper in the cave, the lack of confidence in this and probable irreversibility of the consequences of a wrong decision motivate not to take risks of stumbling in areas with even minor and non-definitive signs of being ``narrow''. It is also reasonable to have a safety gap to the ``walls'' of a ``cave'' when turning inside it. This is possible only if the cave is a bit wider than was just described. So in our characterization of an ``unacceptable cave'', we replace $\rmin$ by a free parameter $\rop$, where
\begin{equation}
\label{hidden.ass}
\rop \in \big(\rmin, d_0 \big) .
\end{equation}
\par
The above precautions are formalized as the requirement to always be in a continuously moving ({\it accompanying}) $\rop$-disc that fully lies in the exterior $\rext$. Any robot's trajectory that meets this requirement is said to be {\it secure}.
\begin{define}
\label{rmin.loop1}
Among the parts into which a Jordan loop
$\mathscr{L} \subset \rext$ cuts the plane, the one without the objects is called the {\em patch} of $\mathscr{L}$.
\end{define}
\begin{define}
\label{rmin.loop2}
A $(R_-,R_+)$-{\em loop} (where $R_\pm >0$) is a regular Jordan loop $\mathscr{L} \subset \rext$
such that
\begin{itemize}
\item $|\kappa_{\mathscr{L}}|\leq 1/R_-$ whenever the signed curvature $\kappa_{\mathscr{L}}<0 $
\item $|\kappa_{\mathscr{L}}|\leq 1/R_+$ whenever the signed curvature $\kappa_{\mathscr{L}} \geq 0$.
\end{itemize}
\end{define}
\par
In Defn.~\ref{rmin.loop2}, the curvature sign corresponds to the loop orientation for which the patch is to the right.
\par
Prop.~\ref{lem.fence}(ii,iii) will point out a set of $(\rop,\rop)$-loops such that any secure trajectory fully lies in the patch of some of them; these loops are called the $\rop$-{\it fences}. Thus the robot cannot approach the targeted equidistant curve $\partial \mathscr{N}^{d_0}(D)$ closer than a $\rop$-fence does.
\par
As a result, the realistic control objective is {\it to arrive at and then track a $\rop$-fence}.
\par
In fact, this is necessarily the $\rop$-fence that encircles the initial location of the robot.
\section{Assumptions and related geometric facts}
\label{sec.ass}
This section discusses general assumptions on the issue of modelling planar work-spaces of robots, as well as certain traits of these models relevant to the main topic of this text.
\par
A curve is said to be {\it analytic} if it is defined by a parametric representation that is locally given by absolutely convergent power series. A {\it segment} of a curve is its arc between two end-points.
\begin{define}
\label{def.pan}
A Jordan loop is said to be {\em piecewise analytic} if it can be represented as the cyclic sequential concatenation of finitely many segments of regular analytic curves.
\end{define}
\par
At the concatenation points, the loop may have kinks.
\par
The following assumption is fulfilled for the overwhelming majority of objects that are of interest in robotics.
\begin{assumption}
\label{ass.obs}
Any set $D_i$ either is a point or is bounded by a piecewise analytic Jordan loop. These sets are disjoint:
 $D_i \cap D_j = \emptyset \; \forall i \neq j$.
\end{assumption}
\par
Here $D_i$ may be either the interior or exterior of a Jordan loop. By the last claim of Asm.~\ref{ass.obs}, there is no more than one ``object'' $D_i$ of the second type.
\par
Any ``non-pointwise'' object is said to be {\it extended}.
We equipp any such object $D_i$ with the {\it natural pa\-ra\-met\-ri\-za\-tion} $\varrho_i(s_i)$ of its boundary $\partial D_i$. Here the {\it curvilinear abscissa} $s_i$ (the arc length) is such that $D_i$ is to the left as $s_i$ ascends and $\varrho_i(s_i) = \varrho_i(s_i\pm p_i)$, where $p_i$ is the perimeter of $\partial D_i$.
Let $\big[ \vec{\tau}_i(s_i) = \frac{d\varrho_i}{d s_i}(s_i), \vec{n}_i(s_i)\big]$ stand for the right-handed Frenet-Serrat frame and $\varkappa_i(s_i)$ for the {\it signed curvature}, i.e., the coefficient in the {\it Frenet-Serrat equations}:
\begin{equation}
\label{fer.serr}
\frac{d\vec{\tau}}{d s_i} = \varkappa_i \vec{n}_i, \quad \frac{d\vec{n}}{d s_i} = - \varkappa_i \vec{\tau}_i.
\end{equation}
Thus $\varkappa_i\geq 0$ for the ``convexities'' of $D_i$, and $\varkappa_i < 0$ for the ``concavities''.
\begin{define}
A {\em bridge} is a straight line segment $S:= [\bldp_1, \bldp_2]$ such that
$[\bldp_1, \bldp_2] \cap D = \{\bldp_1, \bldp_2\}$ and for any of $\bldp_i$'s that lies inside a smooth piece of the boundary of an extended object, $S$ is normal to this boundary at $\bldp_i$.
\end{define}
\begin{define}
\label{def.pliral}
A distance $d_\star >0$ is said to be {\em bizarre} if some of the following statements holds:
\begin{enumerate}[{\bf i)}]
\item There exists an extended object $D_i$ whose boundary contains infinitely many points where $\varkappa_i(s_i\pm) = - d_\star^{-1}$;
\item There are infinitely many locations $\bldr \not \in D$ such that $\dist{\bldr}{D} = d_\star$ and $\pr{\bldr}{D}$ is not a singleton;
\item There exists a bridge whose length equals $2 d_\star$.
\end{enumerate}
\end{define}
\par
Due to Asm.~\ref{ass.obs}, the property from i) means that some concave piece of the boundary $\partial D_i$ is an arc of a $d_\star$-circle.
\begin{prop}
\label{lem.bizarre}
The set of bizarre distances is either finite or empty.
\end{prop}
\par
The proof of this lemma is given in Sec.~\ref{app.2}.
\par
Thus the property of being bizarre is hardly ever encountered (e.g., with the zero probability when drawing $d_\star$ in accordance with a continuous probability distribution) and is destroyed due to arbitrarily small blunders. Meanwhile, analysis becomes much harder if some parameters associated with the problem statement (like $d_0$ or $\rop$)
or the controller (especially, parameters $R$ that have the meaning of a further increased $R \geq \rop$ minimum turning radius $\rmin$) are bizarre. This is an inceptive to focus our analysis on the following case:
\begin{equation}
\label{prereq}
d_0, d_0+\rop, d_0+R \; \text{are not bizarre} \quad \text{and}\quad  R\geq \rop.
\end{equation}
By \eqref{prereq}, the following proposition, in particular, is valid for $R\!:= \!\rop$.
\begin{prop}
Suppose that \eqref{prereq} and Asm.~\ref{ass.union}, \ref{ass.obs} are true. Then the following statements hold:
\begin{enumerate}[{\bf (i)}]
\item  The boundary of any connected component $Q_{\text{cc}}$ of the set
\begin{equation}
\label{def.q}
Q(R) := \{\bldc \in \rext: \dist{\bldc}{\mathscr{P}} > R\}
\end{equation}
is a piecewise analytic Jordan loop;
\item The union of the $R$-discs centered in $Q_{\text{cc}}$ is connected and is edged by a $(R,\rop)$-loop, said to be {\em major};
\item Any secure trajectory lies in the patch of a major $(\rop,\rop)$-loop;
\item Whenever $R \geq R_\ast \geq \rop$, the patch of the major $(R,\rop)$-loop lies in the patch of a uniquely defined major $(R_\ast,\rop)$-loop.
\end{enumerate}
\label{lem.fence}
\end{prop}
\par
The proof of this proposition is given in Sec.~\ref{app.fence}. The items (ii) (with $R:=\rop$) and (iii) of this lemma in fact describe $\rop$-fences (major $(\rop,\rop)$-loops), which should be tracked according to the refined formulation of the control objective, whereas the whole of this lemma provides guidelines for their computation. More precisely,
{\it the task is to track an initial $\rop$-fence}, where a major $(R,\rop)$-loop is called the {\it initial $R$-fence} $\mathscr{F}_R$ if
the initial location $\bldr_{\text{in}}$ lies in its patch.
To make the notion of the initial $R$-fence well-defined, we need one more assumption.
It means that there is enough free space around the initial location of the robot.
\par
Let $C_\pm^{\text{in}}$ stand for the circle travelled by the robot from its initial state under the constant control $\omega \equiv \pm \ov{\omega}$. This is the $\rmin$-circle whose center lies at a distance of $\rmin$ from the initial location $\bldr_{\text{in}}$ on the ray issued from $\bldr_{\text{in}}$ to the left/right perpendicularly to the initial orientation of the robot.
\begin{assumption}
\label{ass.init}
The parameter $R$ is such that $R> R_+:=\dist{\bldr_{\text{\rm in}}}{\mathscr{P}} +3 \rmin$ and the union $C_+^{\text{in}} \cup C_-^{\text{in}}$
fully lies at a distance $<R$ from some connected component $Q_{cc}$ of the set \eqref{def.q}.
\end{assumption}
\par
This tacitly assumes that the set \eqref{def.q} is not empty.
\par
The next fact aids in verifying Asm.~\ref{ass.init} and is proven in Sec.~\ref{app.1}.
\begin{prop}
\label{lem.chch0}
$\dist{\bldr}{\mathscr{P}} = \dist{\bldr}{D} - d_0 \; \forall \bldr \in \rext$.
\end{prop}
\par
In conjunction with other our assumptions, Asm.~\ref{ass.init} does make the significant notion of the initial fence well-defined, as is shown by the following.
\begin{prop}
\label{lem.pere}
Let \eqref{prereq} and Asm.~\ref{ass.union}, \ref{ass.obs}, \ref{ass.init} be true.
Then an initial $R$-fence $\mathscr{F}_R$ and $\rop$-fence $\mathscr{F}_{\rop}$ are well-defined.
\end{prop}
\par
The proof of this proposition is given in Sec.~\ref{app.fence}.

\section{General geometric preliminaries and proof of Proposition~\ref{lem.chch0}.}
\label{app.1}
We consider only planar sets and points from $\br^2$.
\begin{figure}
\centering
\subfigure[]{\scalebox{0.18}{\includegraphics{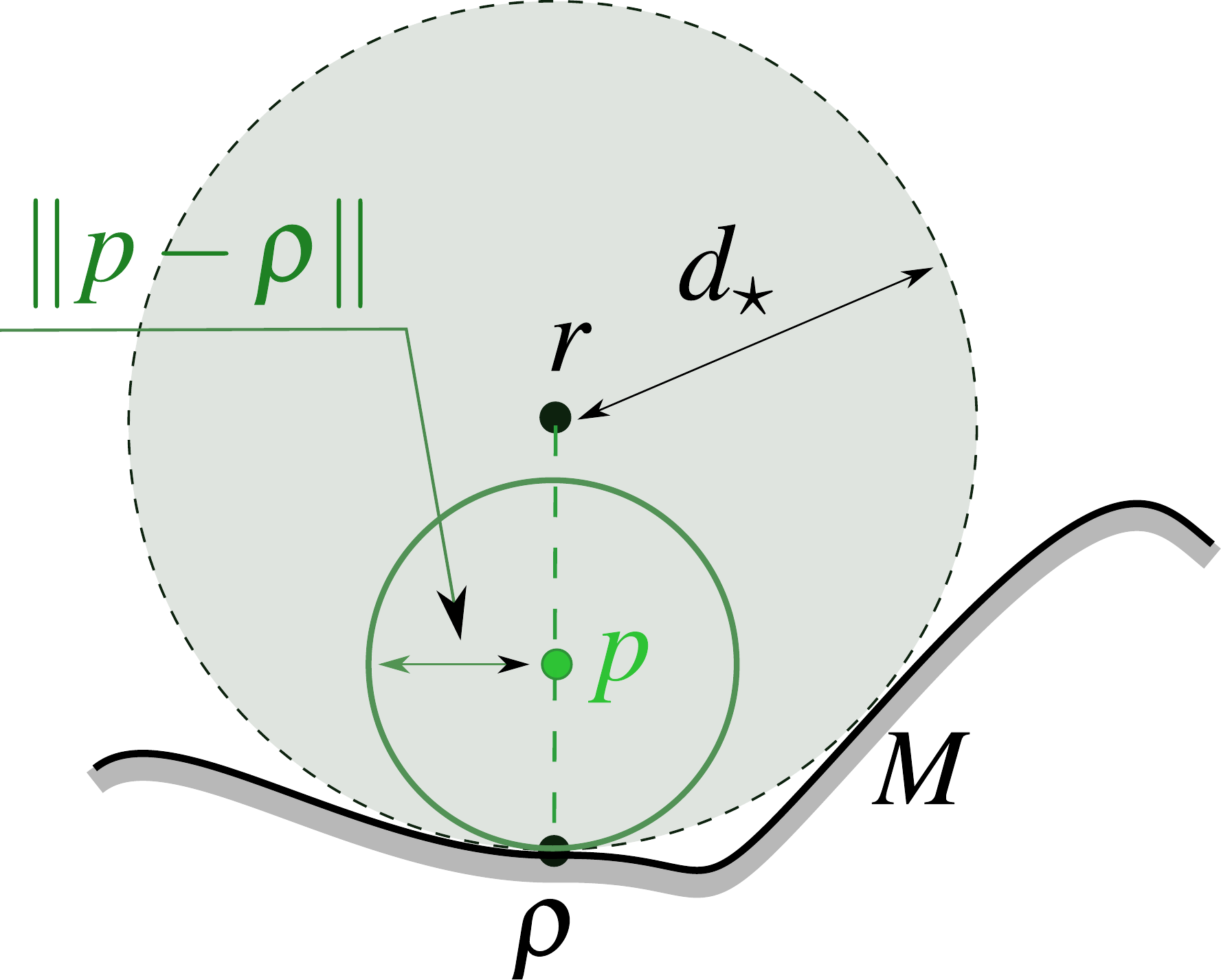}}\label{fig.proj1}}
\caption{Projection onto $M$.}
\label{fig.proj0}
\end{figure}
\begin{lemma}
\label{lem.distt}
Let $M$ be a closed set, $\bldr \not\in M$, and $\rho \in \pr{\bldr}{M}$. Then the following statements hold:
\par
{\bf i)} $\dist{p}{M} = \|p-\rho\|, \pr{p}{M} = \{\rho \}\; \forall p \in [\rho,\bldr)$.
\par
{\bf ii)} The map $p \not\in M \mapsto \pr{p}{M}$ is upper semicontinuous.
\par
{\bf iii)} Let $\bldr(t) \not\in M$ be a $C^1$-function. Then the mapping $z(t) := \dist{\bldr(t)}{M}$ is absolutely continuous and
$\spr{\dot{\bldr}(t)}{N[\bldr(t),\bldp]}$ is independent of $\bldp \in \pr{\bldr(t)}{M}$ and equals $\dot{z}(t)$ for almost all $t$.
\end{lemma}
\begin{IEEEproof}
i) It suffices to note that in Fig.~\ref{fig.proj0}, the interior of the grey disc has no points in common with $M$, whereas the $\|p-\rho\|$-disc centered at $p$ lies in the grey disc, with the only exception of $\rho \in M$.
\par
ii) Suppose that $\rho_\infty = \lim_{k \to \infty} \rho_k$, where $\rho_k \in \pr{p_k}{M}$ and $p_k \not\in M$ for all $k$, and $p_k \to p_\infty \not\in M$ as $k \to \infty$. The proof of ii) is completed by noting that $\rho_\infty \in M$ and, since the function $\dist{\cdot}{M}$ is Lipschitz continuous,
\begin{gather*}
\dist{p_\infty}{M} = \lim_{k \to \infty} \dist{p_k}{M} = \lim_{k \to \infty} \|p_k-\rho_k\|
\\
= \|p_\infty - \rho_\infty\| \Rightarrow \rho_\infty \in \pr{p_\infty}{M}.
\end{gather*}
\par
{\bf iii)} Since the mapping $\bldr \mapsto \dist{\bldr}{M}$ is Lipschitz continuous, so is $t \mapsto z(t)$.
By the Rademacher's theorem, $\dot{z}(t)$ exists for almost all $t$. For any such $t=t_\ast$ and $\bldp \in \pr{\bldr(t_\ast)}{M}$, we have
\begin{gather*}
z(t_\ast)+ \dot{z}(t_\ast)\underbrace{(t-t_\ast)}_{\delta} + \smo(\delta) = z(t) \leq \|\bldr(t) - \bldp\|
\\
 = \underbrace{\|\bldr(t_\ast) - \bldp\|}_{z(t_\ast)}
+ \delta \spr{\dot{\bldr}(t_\ast)}{N[\bldr(t_\ast),\bldp]} + \smo(\delta)
\\
\Rightarrow  \delta \big[ \spr{\dot{\bldr}(t_\ast)}{N[\bldr(t_\ast),\bldp]} - \dot{z}(t_\ast) \big] \geq \smo(\delta).
\end{gather*}
It remains to divide the both sides of the last inequality  by $\delta \neq 0$ and let $\delta \to 0+$ and then $\delta \to 0-$.
\end{IEEEproof}
{\it Proof of Proposition~\ref{lem.chch0}:}
The inclusion $$\mathscr{P} \subset \{p: \dist{p}{D} = d_0\}$$ implies that
\begin{multline}
\rho \in \pr{\bldr}{\mathscr{P}} \Rightarrow \dist{\bldr}{D} \leq \dist{\rho}{D} + \|\bldr - \rho\|
\\
= d_0 + \|\bldr - \rho\| \leq d_0+ \dist{\bldr}{\mathscr{P}}.
\label{hgg}
\end{multline}
For $p_\star \in \pr{\bldr}{D}$ and $p \in \big[ \bldr, p_\star \big]$, i) in Lem.~\ref{lem.distt} states that $\dist{p}{D} = \|p - p_\star\|$, whereas $\|\bldr - p_\star\| =
\dist{\bldr}{D} > d_0$. Hence there exists $p \in \big[ \bldr, p_\star \big]$ such that $\dist{p}{D} = d_0$ and $\dist{\rho}{D} > d_0 \; \forall \rho \in (p,\bldr]$, which means that $p \in \partial \mathscr{N}^{d_0}(D)$ and $p \in \overline{\rext}$, respectively. Hence $p \in \mathscr{P}$ and
\begin{multline}
\dist{\bldr}{\mathscr{P}} \leq \|\bldr - p\| = \|\bldr - p_\star\| - \|p_\star - p\|
\\
 = \dist{\bldr}{D} - d_0. \qquad \IEEEQED
\label{hgg1}
\end{multline}
Given $\bldp \neq \bldr$ and $d>0$, we put
\begin{equation}
\label{def.rp}
N(\bldr,\bldp) := \frac{\bldr-\bldp}{\|\bldr-\bldp\|}, \quad \bldr_{\bldp \vartriangleright d} := \bldp +  d N(\bldr,\bldp).
\end{equation}
\begin{cor}
\label{cor.proj}
The following equation holds:
$$
\pr{\bldr}{\mathscr{P}} = \{ \bldr_{\bldp \vartriangleright d_0} :\bldp \in \pr{\bldr}{D}  \} \quad \forall \bldr  \in \rext.
$$
\end{cor}
\begin{IEEEproof}
By Prop.~\ref{lem.chch0}, the r.h.s. in \eqref{hgg1} equals $\dist{\bldr}{\mathscr{P}}$. So the inequality in \eqref{hgg1} holds as equality, $p \in \pr{\bldr}{\mathscr{P}}$, and
$\{ \bldr_{\bldp \vartriangleright d_0} : \bldp \in \pr{\bldr}{D}  \} \subset \pr{\bldr}{\mathscr{P}}$.
In \eqref{hgg}, the inequalities similarly hold as equalities. For $\varrho \in \pr{\rho}{D}$, it follows that
\begin{multline*} \|\bldr - \rho\| + \|\rho - \varrho\| = \dist{\bldr}{D} \leq \|\bldr - \varrho\| \Rightarrow \rho \in [\varrho, \bldr].
\end{multline*}
Since $\|\rho - \varrho\| = d_0$, we have $\{ \bldr_{\bldp \vartriangleright d_0} : \bldp \in \pr{\bldr}{D}  \} \supset \pr{\bldr}{\mathscr{P}}$, which completes the proof.
\end{IEEEproof}
\begin{lemma}
\label{lem.anal}
Let $\gamma_1, \gamma_2 \subset \br^2$ be segments of analytical curves. Then one of the following statements holds:
\begin{enumerate}[{\bf i)}]
\item $\gamma_1$ and $\gamma_2$ have only finitely many points in common;
\item $\gamma_1$ and $\gamma_2$ are hosted by a common analytical curve and  $\gamma_1 \cap \gamma_2$ is its segment.
\end{enumerate}
\end{lemma}
\begin{IEEEproof}
Let i) fail to be true. Then there exists an infinite sequence $ \{ \bldp_k \}_{k=1}^\infty \subset \gamma_1 \cap \gamma_2$ of pairwise distinct points.
By passing to a subsequence, we ensure that there exists the limit $\bldp = \lim_{k \to \infty} \bldp_k \in \gamma_1 \cap \gamma_2$. Let
\begin{equation}
\label{exp.ser}
q_i(\varsigma_i) = \bldp + \sum_{m=1}^\infty \vec{a}_{i,m}\varsigma_i^m, \quad \varsigma_i \approx 0
\end{equation}
be an analytical parametrization of $\gamma_i$ in a vicinity of $\bldp$, and let $m_i$ be the minimal index $m$ for which $\vec{a}_{i,m} \neq 0$. By passing to a subsequence once more and multiplying the odd coefficients in \eqref{exp.ser} by $-1$ (if necessary), we ensure that $\varsigma_i>0$ for all $\bldp_k$'s. If $\vec{a}_{1,m_1}$ and $\vec{a}_{2,m_2}$ do not lie on a common ray, then for $i=1,2$ and $\varsigma_i \approx 0, \varsigma_i > 0$, the functions \eqref{exp.ser} (with $i=1,2$) assume values in disjoint small angles centered at $\vec{a}_{1,m_1}$ and $\vec{a}_{2,m_2}$, respectively, which is
inconsistent with existence of infinitely many common values. Hence $\vec{a}_{1,m_1} = c \cdot \vec{a}_{2,m_2}$, where $c>0$. Then via a proper change of the variable $\varsigma_1$, we can ensure that $\vec{a}_{1,m_1}=\vec{a}_{2,m_2}$. Let $\vec{a}$ stand for this common value.
\par
We consider the Cartesian frame whose center is at $\bldp$ and the abscissa axis goes in the direction of $\vec{a}$. Due to \eqref{exp.ser}, the abscissa $x_i(\varsigma_i)$ of $q_i(\varsigma_i)$ has the form $x_i(\varsigma_i) = \varsigma_i^{m_i} f_i(\varsigma_i)$, where $f_i(\cdot)$ is an analytic function and $f_i(0) \neq 0$. In the equation $x = x_i(\varsigma_i) \Leftrightarrow x^{1/m_i} = g_i(\varsigma_i):=\varsigma_i f_i(\varsigma_i)^{1/m_i}$, the r.h.s. is analytic near $0$ and $g^\prime_i(0) \neq 0$. So
the inverse to $g_i(\cdot)$ is an analytic function and hence its positive root $\varsigma_i(x)$ expands into Puisaux series near zero:
$\varsigma_i(x) = \sum_{m=1}^\infty b_{i,m} x^{\frac{m}{m_i}}$. Then the change of the variable $\varsigma_i \mapsto x$ transforms \eqref{exp.ser} into Puisaux series with respect to the parameter $x$:
\begin{equation}
\label{exp.ser1}
\widetilde{q}_i(x) = \bldp + \sum_{m=1}^\infty \vec{g}_{i,m} x^{\frac{m}{m_i}} .
\end{equation}
This parameter $x$ is common for $i=1,2$.
Let $\mathfrak{x}_k$ be the abscissa of $\bldp_k$. For $x:= \mathfrak{x}_k$, the l.h.s. of \eqref{exp.ser1} assumes a common value for $i=1,2$. By equating the respective r.h.s.'s, we see that in \eqref{exp.ser1} with $i=1$ and $i=2$, respectively, the terms of the least degree are common.
After dropping them on the left and right, the same argument shows that the terms with the next subsequent minimal degree are also common. By continuing likewise, we establish the identity of the series in \eqref{exp.ser1}.
\par
Thus $\gamma_1 \cap \gamma_2$ contains a segment of an analytical curve. Let us maximally extend this segment, while remaining inside $\gamma_1 \cap \gamma_2$. By the foregoing, any end-point of this extended segment is an end of either $\gamma_1$ or $\gamma_2$. This implies ii) and completes the proof.
\end{IEEEproof}
\par
The {\it interior} $\overset{\circ}{\gamma}$ of the curve's segment $\gamma$ is this segment $\gamma$ minus its ends.
\begin{lemma}
\label{for.norra}
Suppose that $\mathscr{L}$ is a  piecewise analytic Jordan loop, $\mathscr{S}$ is one of the two parts into which $\mathscr{L}$ splits the plane, $\mathscr{L}$ is oriented so that $\mathscr{S}$ is to the right, $\big[ \vec{\tau}_{\mathscr{L}}(\rho\pm), \vec{n}_{\mathscr{L}}(\rho\pm)\big]$ and $\varkappa_{\mathscr{L}}(\rho \pm)$ are the right-handed Frenet-Serrat frame and the
signed curvature of the loop $\mathscr{L}$ at its point $\rho$.
Let $\bldr \in \mathscr{S}$ and $\rho \in \pr{\bldr}{\mathscr{L}}$, and let $d_\star$ stand for $\dist{\bldr}{\mathscr{L}}>0$.
\par
 Then the following statements hold:
\begin{enumerate}[{\bf i)}]
\item the unit tangent vector $\vec{\tau}_{\mathscr{L}}(\rho+)$ results from $\vec{\tau}_{\mathscr{L}}(\rho-)$ via a counterclockwise rotation through an angle $\alpha \leq \pi$, and $\bldr = \rho - d_\star \vec{n}$, where $\|\vec{n}\| =1$ and
\begin{enumerate}
\item if $\alpha < \pi$, then there exist $c_\pm \geq 0$ such that
\begin{equation}
\label{decom.nneg}
\vec{n} = c_-\vec{n}_{\mathscr{L}}(\rho-)+c_+\vec{n}_{\mathscr{L}}(\rho+);
\end{equation}
\item if $\alpha = \pi$, then $$\vec{n} \in \{\vec{w}: - \spr{\vec{w}}{\vec{\tau}_{\mathscr{L}}(\rho-)} = \spr{\vec{w}}{\vec{\tau}_{\mathscr{L}}(\rho+)} \geq 0\};$$
\end{enumerate}
\item If $\vec{n} = \vec{n}_{\mathscr{L}}(\rho+)$, then $1+d_\star \varkappa_{\mathscr{L}}(\rho+) \geq 0$ and $\rho$ is immediately followed by a piece $\mathpzc{p}$ of $\mathscr{L}$ such that
    \begin{itemize}
    \item either {\bf a)} $\mathpzc{p}$ is an arc of a $d_\star$-circle with negative curvature or {\bf b)} $1+d_\star \varkappa_{\mathscr{L}}(\widetilde{\rho}) > 0\; \forall \widetilde{\rho} \in \overset{\circ}{\mathpzc{p}}$;
        \end{itemize}
\item If $\vec{n} = \vec{n}_{\mathscr{L}}(\rho-)$, then $1+d_\star \varkappa_{\mathscr{L}}(\rho-) \geq 0$ and $\rho$ is immediately preceded by a piece $\mathpzc{p}$ of $\mathscr{L}$ such that either ii.a) or ii.b) is true;
\item Let $\mathpzc{p}$ be the segment from either ii) or iii) and let $\mathpzc{p}$ has the property b).
Then this segment can be reduced (if necessary) so that $\rho$ is still its end and for any $\widetilde{\rho} \in \overset{\circ}{\mathpzc{p}}$ and $d \in (0,d_\star]$, the $d$-circle centered at $\widetilde{\rho} - d\vec{n}_{\mathscr{L}}(\widetilde{\rho})$ and $\mathpzc{p}$ have only one common point $\widetilde{\rho}$.
\end{enumerate}
\end{lemma}
\begin{IEEEproof}
{\bf i)} We consider separately two cases.
\par
$\boxed{\boldsymbol{\vec{\tau}_{\mathscr{L}}(\rho+) \neq - \vec{\tau}_{\mathscr{L}}(\rho-)}}$. Then $\vec{n}_{\mathscr{L}}(\rho+) \neq - \vec{n}_{\mathscr{L}}(\rho-)$
and so $\mathfrak{N}:= \big\{ \vec{\mathfrak{n}}: \spr{\vec{\mathfrak{n}}}{\vec{n}_i(\rho\pm)} > 0 \big\} \neq \emptyset$.
For $\vec{\mathfrak{n}} \in \mathfrak{N}$, we have $\rho + t \vec{\mathfrak{n}} \not\in \mathscr{S} \; \forall t>0, t \approx 0$.
Hence the function $t>0 \mapsto f(t):= \| \rho + t \vec{\mathfrak{n}} - \bldr\|^2$ attains its local minimum at $t=0$ and so
$0 \leq f^\prime(0) = 2\spr{\vec{\mathfrak{n}}}{\rho- \bldr}$.
By continuity, this inequality extends on the closed cone $\ov{\mathfrak{N}}:= \big\{ \vec{\mathfrak{n}}: \spr{\vec{\mathfrak{n}}}{\vec{n}_i(\rho\pm)} \geq 0 \big\}$ dual to the set $\{ \vec{n}_i(\rho-), \vec{n}_i(\rho+) \}$ and means that $\rho - \bldr $ is in the cone $\ov{\mathfrak{N}}^\ast$ dual to $\ov{\mathfrak{N}}$.
By the bipolar theorem, $\ov{\mathfrak{N}}^\ast$ consists of nonnegative linear combinations of $\vec{n}_i(\rho-)$ and $\vec{n}_i(\rho+)$, whereas $\|\bldr-\rho\| =d_\star$. Hence $\bldr = \rho -d_\star \vec{n}$, where $\vec{n} := d^{-1}_\star(\rho - \bldr)$ satisfies \eqref{decom.nneg} with some $c_\pm \geq 0$.
\par
Let i) fail to be true. Then $\vec{\tau}(\rho+)$ results from $\vec{\tau}(\rho-)$ via a clockwise rotation through a nonzero angle $< \pi$. Then $\rho + t \vec{n} \not\in \mathscr{S} \; \forall t>0, t \approx 0$ for any $\vec{n}\neq 0$ from the open reflex angle $A$ swept as the ray hosting $-\vec{\tau}(\rho-)$ rotates clockwise until arrival at $\vec{\tau}(\rho+)$. By retracing the arguments from the first paragraph, we see that $\spr{\vec{n}}{\rho -\bldr } \geq 0$ for any $\vec{n} \in A$. This implies that $d_\star=\|\bldr-\rho\| =0$, in violation of $d_\star>0$. This contradiction completes the proof of i).
\par
$\boxed{\boldsymbol{\vec{\tau}_{\mathscr{L}}(\rho+) = - \vec{\tau}_{\mathscr{L}}(\rho-)}}$. Let $\varrho_+(s), s \geq  0, s \approx 0$ be the natural parametric representation of a small piece of $\mathscr{L}$ following $\rho = \varrho(0)$. Then $\vec{\tau}_{\mathscr{L}}(\rho+) = \frac{\varrho(s)}{ds}(0+) $ and $g(s) := \frac{1}{2} \|\bldr - \varrho(s)\|^2 \geq g(0)$. So
\begin{gather*}
0 \leq g^\prime(0+) =  \spr{\rho -\bldr}{\vec{\tau}_{\mathscr{L}}(\rho+)} \Rightarrow \spr{\vec{n}}{\vec{\tau}_{\mathscr{L}}(\rho+)} \geq 0
\end{gather*}
for $\vec{n} := d^{-1}_\star(\rho - \bldr)$, which completes the proof of i).
\par
{\bf ii)} We note that $g(s) := \frac{1}{2}\|\varrho(s) - \bldr\|^2 \geq g(0)$ and $g^\prime(s) = \spr{\varrho(s) - \bldr}{\vec{\tau}_{\mathscr{L}}\big[\varrho(s)\big]}$.
By \eqref{fer.serr} and i),
\begin{multline*}
g^\prime(0+)= \spr{\rho - \bldr}{\vec{\tau}_{\mathscr{L}} (\rho+)} = d_\star^{-1} \spr{\vec{n}_{\mathscr{L}} (\rho+)}{\vec{\tau}_{\mathscr{L}} (\rho+)} =0
\\
\Rightarrow 0 \leq g^{\prime\prime}(0+)
\\
= \spr{\vec{\tau}_{\mathscr{L}}(\rho+)}{\vec{\tau}_{\mathscr{L}}(\rho+)} + \varkappa_{\mathscr{L}}(\rho+)\spr{\rho-\bldr}{\vec{n}_{\mathscr{L}}(\rho+)}
\\
= 1 + d_\star \varkappa_{\mathscr{L}}(\rho+).
\end{multline*}
\par
Thus $\varkappa_{\mathscr{L}}(\rho+) \geq -d_\star^{-1}$. Let $\mathpzc{p}$ be an analytic segment of $\mathpzc{L}$ that immediately follows $\rho$. Since the function $\widetilde{\rho} \in \mathpzc{p} \mapsto \varkappa_{\mathscr{L}}(\widetilde{\rho}) + d_\star^{-1}$ is analytic it is either 1) identically $0$ or 2) has no more than finitely many roots. In the case 1), a) in ii) holds. Let the case 2) occur. By reducing $\mathpzc{p}$ (if necessary), we ensure that $\varkappa_{\mathscr{L}}(\widetilde{\rho})  \neq - d_\star^{-1} \; \forall \widetilde{\rho} \in \overset{\circ}{\mathpzc{p}}$ and $\rho \in \mathpzc{p}$.
\par
Suppose that b) in ii) does not hold. Then $\varkappa_{\mathscr{L}}(\widetilde{\rho})  < - d_\star^{-1} \; \forall \widetilde{\rho} \in \overset{\circ}{\mathpzc{p}}$.
We introduce the Cartesian frame centered at $\rho$ whose ordinate axis is aimed at $\bldr$ and the abscissa axis is co-directed with $\mathscr{L}$.
Near the origin, $\mathpzc{p}$ is the graph of a function $y=h(x), x \in [0,x_+], x_+>0$.
Hence $\varkappa_{\mathscr{L}} = \frac{-h^{\prime\prime}}{(1+(h^\prime)^2)^{3/2}} <- d_\star^{-1}$ and so $q^{\prime} > d_\star^{-1} (1+q^2)^{3/2}$ for $q:= h^\prime$ and $x \in (0,x_+]$, where $q(0) =0$. By \cite[Ch.~III, Th.~4.1]{Hart82}, $[q(x) - q_\ast(x)] >0\; \forall x > 0, x \approx 0$, where $q_\ast(\cdot)$ is the solution of the Cauchy problem $ q_\ast(0) =0$ for ODE $q^{\prime}_\ast = d_\star^{-1} (1+q_\ast^2)^{3/2}$. So $q_\ast(x) = \frac{x}{\sqrt{d_\star^2-x^2}}$.
Hence
\begin{multline*}
h(x) = \int_0^x q(z) \; dz
\\ > \int_0^x q_\ast(z) \; dz = d_\star^2-\sqrt{d_\star^2-x^2} \quad \forall x > 0, x \approx 0.
\end{multline*}
Hence $\mathpzc{p} \subset \mathscr{L}$ penetrates inside the $d_\star$-disc centered at $\bldr$ and so $\dist{\bldr}{\mathscr{L}} < d_\star$, in violation of the definition of $d_\star$. This contradiction completes the proof of b).
\par
{\bf iii)} is proved likewise.
\par
{\bf iv)} Let $\mathpzc{p}$ be the segment from ii); the case of iii)
is handled likewise. It suffices to examine $d \in [d_\star/2, d_\star]$.
Now $q^{\prime} < d^{-1} (1+q^2)^{3/2}\; \forall x \in (0,x_+]$. Let $\widetilde{x}$ stand for the abscissa of point $\widetilde{\rho} \in \mathpzc{p}$.
By \cite[Ch.~III, Th.~4.1]{Hart82},
\begin{equation*}
[q(x) - q_\ast(x)] \sgn (x-\widetilde{x}) < 0\; \forall x \in [0,x_+]\cap \mathscr{D}^{q_\ast}, x \neq \widetilde{x},
\end{equation*}
where $q_\ast(\cdot)$ is the solution of the Cauchy problem
$$
q^{\prime}_\ast = d^{-1} (1+q_\ast^2)^{3/2}, \qquad q_\ast(\widetilde{x}) = q(\widetilde{x})
$$
and $\mathscr{D}^{q_\ast}$ is the domain of definition of $q_\ast(\cdot)$.
Here
\begin{gather*}
q_\ast(x) = \frac{x-c}{\sqrt{d^2-(x-c)^2}}, \quad c : = \widetilde{x} - \frac{d q(\widetilde{x})}{\sqrt{1+ q(\widetilde{x})^2}},
\\
\mathscr{D}^{q_\ast} = \big\{ x: |x-c| < d \big\}.
\end{gather*}
\par
For $x \in [0,x_+]\cap \mathscr{D}^{q_\ast}, x \neq \widetilde{x}$, we have
\begin{gather*}
h(x) = h(\widetilde{x})+\int_{\widetilde{x}}^x q(z) \; dz <  h(\widetilde{x})+\int_{\widetilde{x}}^x q_\ast(z) \; dz
\\
= h(\widetilde{x})+ \sqrt{d^2-(\widetilde{x}-c)^2} -\sqrt{d^2-(x-c)^2}.
\end{gather*}
The graph of the function on the right is the $d$-circle that passes through the point $(\widetilde{x},h(\widetilde{x}))$ and is tangential to the graph of $h(\cdot)$ at this point. Simultaneously, this is nothing but the $d$-circle $\mathscr{C}$ addressed in iv), and $\mathscr{D}^{q_\ast}$ is its projection onto the abscissa axis. Thus we see that among the points of the plane whose abscissas lie in this projection, the circle $\mathscr{C}$ has no points in common with $\mathpzc{p}$, except for $\w{\rho}$. It remains to note that $\mathscr{C} \cap \mathpzc{p}$ lies amidst those points.
\end{IEEEproof}
\begin{lemma}
\label{lem.arc}
Let the assumptions of Lem.~\ref{for.norra} hold, $\bldr$ and $\rho$ are taken from Lem.~\ref{for.norra}, whereas $\mathpzc{p}$ is taken from ii) or iii) in Lem.~\ref{for.norra}.
Suppose that $\vec{n}_{\mathscr{L}}(\rho-)=\vec{n}_{\mathscr{L}}(\rho+)=: \vec{n}$ and $\bldr(t) \in \mathscr{S}, t \geq t_0$ is a $C^k$-smooth ($k \geq 1$) trajectory such that $\bldr(t_0) = \bldr$, $\spr{\dot{\bldr}(t_0)}{\vec{n}} >0$, and the tangential part of $\dot{\bldr}(t_0)$ looks in the direction of $\mathpzc{p}$.
\par
Then for $t>t_0, t \approx t_0$, the projection $\pr{\bldr(t)}{\mathpzc{p}}$ contains a single point $\rho(t)$. This point lies in $\overset{\circ}{\mathpzc{p}}$, is a
$C^k$-function of $t$, and $1+\md(t) \varkappa_{\mathscr{L}}[\rho(t)] >0$, where $\md(t):= \dist{\bldr(t)}{\mathpzc{p}}$.
\end{lemma}
\begin{IEEEproof}
If the case {\bf a)} from ii) in Lem.~\ref{for.norra} holds, the proof is immediate from Fig.~\ref{fig.proj1}.
\begin{figure}
\centering
\subfigure{\scalebox{0.17}{\includegraphics{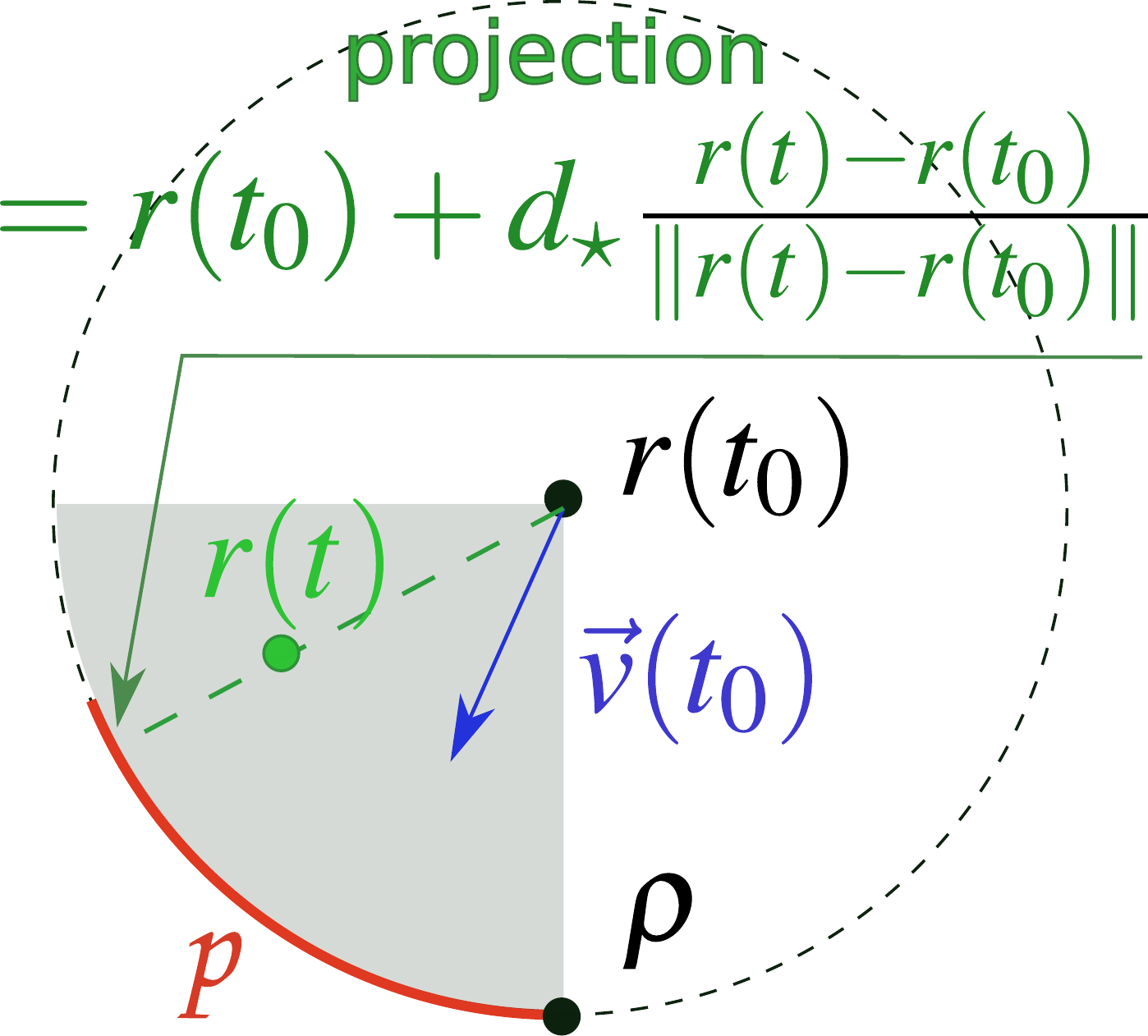}}\label{fig.proj21}}
\caption{Projection onto an arc of a circle}
\label{fig.proj1}
\end{figure}
\par
Suppose that the case {\bf b)} holds.
We reduce the segment $\mathpzc{p}$ according to iv) in Lem.~\ref{for.norra}, form $\w{\mzp}$
by complementing $\mzp$ with a short straight segment that tangentially goes from $\rho$ away from $\mathpzc{p}$,
put $\md_\star:=\md(t_0)$, and focus on $t>t_0, t \approx t_0$.
Thanks to
iv) in Lem.~\ref{for.norra}, $\pr{\bldr}{\w{\mathpzc{p}}} = \{\rho\}$ and so
$\w{\rho} \to \rho$ for $\w{\rho} \in \pr{\bldr(t)}{\w{\mathpzc{p}}}$ as $t \to t_0+$ by ii) in Lem.~\ref{lem.distt}.
By i) in  Lem.~\ref{for.norra}, $\bldr(t) = \w{\rho} - \md(t) \vec{\mzn}[\w{\rho}]$, where $\vec{\mzn}$ is the unit normal to $\mathpzc{p}, \mzn(\rho)= \vec{n}$. If $\w{\rho} \in \pr{\bldr(t)}{\w{\mathpzc{p}}} \setminus \mzp$, then
\begin{gather*}
\w{\rho} - \rho = \bldr(t)- \bldr(t_0)+ \md(t) \big\{ \underbrace{\vec{\mzn}[\w{\rho}] - \vec{n}}_{=0}\big\} - [\md_\star -\md(t) ] \vec{n}
\\
= \vec{v}(t_0) (t-t_0) + \so (t-t_0)  - [\md_\star -\md(t) ] \vec{n};
\\
\spr{\w{\rho} - \rho}{\vec{\tau}_{\mathscr{L}}(\rho)} = \spr{\vec{v}(t_0)}{\vec{\tau}_{\mathscr{L}}(\rho)} (t-t_0) + \so (t-t_0) .
\end{gather*}
Since the tangential component of $\vec{v}(t_0)$ looks at $\mzp$, so thereby is $\w{\rho} - \rho$. Hence $\w{\rho} \in \mzp$ for $t \approx t_0$, in violation of $\w{\rho} \not\in \mzp$. Thus we see that $\pr{\bldr(t)}{\w{\mathpzc{p}}}\setminus \mzp = \emptyset$ and so $$\pr{\bldr(t)}{\w{\mathpzc{p}}} = \pr{\bldr(t)}{\mathpzc{p}}\subset \overset{\circ}{\mzp}.$$
\par
\par
{\bf 1)} Suppose first that $1+\md_\star \varkappa_{\mathscr{L}}(\rho) >0$. Let $\varrho(s), s \geq 0, \varrho(0) = \rho$  be the natural parametric representation of $\mzp$.
For $f(s,d) := \varrho(s) - d \vec{n}_{\mathscr{L}}[\varrho(s)]$, we have
$f^\prime_s = (1+\varkappa_{\mathscr{L}} d) \vec{\tau}_{\mathscr{L}}, f^\prime_d = - \vec{n}_{\mathscr{L}}$. So
the Jacobian determinant of $f(\cdot)$ at $s=0, d=\md_\star$ is nonzero, and the implicit function  theorem \cite[Thm.~21.11]{bartle1976} yields that
for $t \approx t_0, t >t_0$, the equation $f[s,d] = \bldr(t)$ has a unique solution $s=s(t), d=d(t)$, which is a $C^k$-function of $t$.
It remains to note that the coordinate $s$ of any $\w{\rho} \in \pr{\bldr(t)}{\mathpzc{p}}$ and $d := \md(t)$ solve this equation.
\par
{\bf 2)} It remains to examine the case where $1+\md_\star \varkappa_{\mathscr{L}}(\rho) =0$.
Since $\frac{d}{dt} \|\bldr(t) - \rho\|^2|_{t=t_0+} <0$, we have $\md(t) \leq \|\bldr(t) - \rho\| < \md_\star$.
Then iv) in Lem.~\ref{for.norra} implies that $\pr{\bldr(t)}{\mathpzc{p}}$ contains a single point $\rho(t)$.
By repeating the arguments from 1) near any of these points, we see that the map $t>t_0, t \approx t_0 \mapsto \rho(t)$ is $C^k$ smooth.
\end{IEEEproof}

\section{Proof of Proposition~\ref{lem.bizarre}.}
\label{app.2}
Asm.~\ref{ass.obs} is adopted from now on. Then by Defn.~\ref{def.pan}, the boundary $\partial D_i$ of any extended object is a piecewise analytic Jordan loop. Let
$\gamma_1^{i}, \ldots, \gamma_{k(i)}^i$ be an associated cyclic sequence of segments from  Defn.~\ref{def.pan}.
\begin{lemma}
There are no more than finitely many values of $d_\star$ for which i) in Defn.~\ref{def.pliral} holds.
\label{i.fm}
\end{lemma}
\begin{IEEEproof}
For any such $d_\star$, some piece $\gamma^i_j$ hosts infinitely many points with $\varkappa_i(s_i) = - d_\star^{-1}$. Since $\varkappa_i(\cdot)$ is analytic, it is constant and so $\gamma^i_j$ is an arc of a $d_\star$-circle. The total number of these circles and their radii do not exceed the number of $\gamma^p_q$'s and so is finite.
\end{IEEEproof}
\par
For any segment $\gamma^i_j$ and number $d_\star>0$, let $\gamma_j^i(d_\star)$ be the segment parametrically given by
$$
\gamma^i_j \mapsto \varrho_i(s_i|d_\star):=\varrho_i(s_i) - d_\star \vec{n}_i(s_i).
$$
For any pointwise object $D_i = \{\bldp_i\}$, we split the $d_\star$-circle centered at the point $\bldp_i$ into cyclically concatenated arcs $\gamma^i_1(d_\star), \ldots, \gamma^i_{k_i}(d_\star)$ and parametrize them so that $\bldp_i$ is to the left as the parameter ascends. For any extended object $D_k$, we do the same for any kink point $\bldp_k$ of $\partial D_k$ where the first sentence from i) in Lem.~\ref{for.norra} is valid. The respective segments $\gamma_j^i(d_\star)$ are numbered by $i$ outside the index range of the objects. Some segments $\gamma_j^i(d_\star)$ may degenerate into a point; this holds if $\gamma_i^j$ is a circular segment with the negative curvature $-\frac{1}{d_\star}$.
\begin{define}
\label{ensemble}
The so obtained collection of the segments $\gamma_j^i(d_\star)$ is called the $d_\star$-{\em ensemble}, and the associated set of the ``parent'' segments $\gamma_j^i$'s its {\em base}, if
\begin{enumerate}[{\bf s.1)}]
\item  whenever $D_i$ is an extended object, either $1+d_\star \varkappa_i(s_i) \neq 0$ everywhere on $\overset{\circ}{\gamma}\vphantom{\gamma}^i_j$ or $1+d_\star \varkappa_i(s_i) \equiv 0$
everywhere on $\overset{\circ}{\gamma}\vphantom{\gamma}^i_j$;
\item $\overset{\circ}{\gamma}\vphantom{\gamma}_j^i(d_\star)$ and $\overset{\circ}{\gamma}\vphantom{\gamma}_q^p(d_\star)$ are disjoint or identical whenever $(i,j) \neq (p,q)$;
\item no segment $\gamma^i_j(d_\star)$ intersects itself.
\end{enumerate}
\end{define}
\par
Lem.~\ref{lem.anal} and further partitioning the segments (if necessary) show that $d_\star$-ensembles do exist.
\begin{lemma}
Suppose that $1+d_\star \varkappa_i(s_i)  > 0$. Then the following statements hold:
\begin{enumerate}[{\bf i)}]
\item $ [\vec{\tau}_i(s_i), \vec{n}_i(s_i)]$ is the Frenet-Serret frame of $\gamma_j^i(d_\star)$ at the point $\varrho_i(s_i|d_\star)$ provided that $\gamma_j^i(d_\star)$ is oriented in the ascending order of $s_i$;
\item the curvature of the segment $\gamma_j^i(d_\star)$ at the point $\varrho_i(s_i|d_\star), s_i \in \overset{\circ}{\gamma}\vphantom{\gamma}^i_j$ equals $\frac{\varkappa_i(s_i)}{1+d_\star \varkappa_i(s_i)}$.
\end{enumerate}
\label{lem.curv}
\end{lemma}
\begin{IEEEproof}
By \eqref{fer.serr},
$$
\frac{\varrho_i(s_i|d_\star)}{d s_i} = \big[1+d_\star \varkappa_i(s_i) \big]\vec{\tau}_i(s_i).
$$
So the claim {\bf i)} does hold and
the differentials of the arc lengths $\widetilde{s}_i$ and $s_i$ over $\gamma_j^i(d_\star)$ and $\gamma_j^i$, respectively, are linked by
$d \widetilde{s}_i = \big[1+d_\star \varkappa_i(s_i) \big] \,ds_i$. The proof is completed by \eqref{fer.serr} applied to the segment $\gamma_j^i(d_\star)$.
\end{IEEEproof}
\begin{lemma}
There are no more than finitely many values of the distance $d_\star$ for which ii) in Defn.~\ref{def.pliral} holds.
\label{i.fm1}
\end{lemma}
\begin{IEEEproof}
Thanks to Lem.~\ref{i.fm}, it suffices to show that there are no more than finitely many values of $d_\star$ for which ii) in Defn.~\ref{def.pliral} holds but i) is not true.
Let $d_\star$ be such value. Then  there is an infinite set $\mathfrak{R}$ of locations $\bldr \not \in D$ such that $\dist{\bldr}{D} = d_\star$ and $\pr{\bldr}{D}$ contains points $\rho_1(\bldr) \neq \rho_2(\bldr)$. Any of them belongs to either a pointwise object or to a segment $\gamma_j^i$. Properly thinning out the set $\mathfrak{R}$ results in a situation where for any $\nu=1,2$, either $\rho_\nu(\bldr)$'s are common for all $\bldr \in \mathfrak{R}$ or for all $\bldr \in \mathfrak{R}$, they are pair-wise distinct and lie on $\overset{\circ}{\gamma}\vphantom{\gamma}_j^i$ for a  common $(i,j)$. We shall consider separately three cases.
\par
{\bf Case 1:} Both $\rho_1:=\rho_1(\bldr)$ and $\rho_2:=\rho_2(\bldr)$ do not depend on $\bldr$. Then all $\bldr$'s lie at the intersection of two $d_\star$-circles that are centered at $\rho_1$ and $\rho_2$, respectively. These circles have no more than two points in common, whereas the set $\mathfrak{R}$ of $\bldr$'s is infinite. Thus the case 1 is impossible.
\par
{\bf Case 2:} One of the collections $\{\rho_\nu(\bldr) \}_{\bldr \in \mathfrak{R}}, \nu =1,2$ consists of a single element (say $\rho_1(\bldr)=: \rho_1$), whereas the elements of the other collection are pairwise distinct and lie on a common segment $\gamma_j^i$.
Since $d_\star$ does not meet i) in Defn.~\ref{def.pliral}, the segment $\gamma^i_j(d_\star)$ does not degenerate into a point.
Meanwhile,
$\mathfrak{R}$ is a subset of both $\gamma^i_j(d_\star)$ and the $d_\star$-circle centered at $\rho_1$.
Then $\gamma^i_j(d_\star)$ lies on this circle by Lem.~\ref{lem.anal}.
and so its curvature is $- d_\star^{-1}$ everywhere.
By ii,iii) in Lem.~\ref{for.norra}, $1+d_\star \varkappa_i(s_i) >0$ on $\overset{\circ}{\gamma}\vphantom{\gamma}^i_j$.
Then Lem.~\ref{lem.curv} implies that $\varkappa_i(s_i) = - \frac{1}{2 d_\star}$ on $\gamma_j^i$.
Thus $\gamma^i_j$ is a circular arc and $d_\star$ is half its radius.
Since among $\gamma_q^p$'s, the number of the circular arcs is finite, the case 2 holds for no more than finitely many $d_\star$'s.
\par{\bf Case 3:} The elements of the collection $\{\rho_\nu(\bldr) \}_{\bldr \in \mathfrak{R}}$ are pair-wise distinct for $\nu =1,2$.
There exist $(i,j)$ and $(p,q)$ such that $\rho_1(\bldr) \in \overset{\circ}{\gamma}\vphantom{\gamma}^i_j, \rho_2(\bldr) \in \overset{\circ}{\gamma}\vphantom{\gamma}^p_q \; \forall \bldr \in \mathfrak{R}$.
By ii,iii) in Lem.~\ref{for.norra}, $1+d_\star \varkappa_i(s_i) >0$ on $\overset{\circ}{\gamma}\vphantom{\gamma}^i_j$ and $1+d_\star \varkappa_p(s_p) >0$ on $\overset{\circ}{\gamma}\vphantom{\gamma}^p_q $.
The segments $\gamma^i_j(d_\star)$ and $\gamma^p_q(d_\star)$ have infinitely many points in common. By Lem.~\ref{lem.anal}, they lie on a common curve $C$.
Let $s_i(\bldr)$ and $s_p(\bldr)$ be the coordinates of $\rho_1(\bldr)$ and $\rho_2(\bldr)$, respectively, and let $\gamma_j^i(d_\star)$ and $\gamma_q^p(d_\star)$ be oriented like in i) of Lem.~\ref{lem.curv}. If these orientations are consistent, then $\Big\{\vec{\tau}_i\big[s_i(\bldr) \big], \vec{n}_i\big[s_i(\bldr)\big] \Big\}$ and
$\Big\{\vec{\tau}_p\big[s_p(\bldr) \big], \vec{n}_p\big[s_p(\bldr)\big] \Big\}$ are the Frene-Serret frames of $C$ at $\bldr$ by {\bf i)} in Lem.~\ref{lem.curv}. Then $\vec{n}_i\big[s_i(\bldr)\big] = \vec{n}_p\big[s_p(\bldr)\big]=: \vec{n}$ and
$$
\rho_1(\bldr) - d_\ast \vec{n} = \bldr = \rho_2(\bldr) - d_\ast \vec{n} \Rightarrow \rho_1(\bldr) = \rho_2(\bldr),
$$
in violation of $\rho_1(\bldr) \neq \rho_2(\bldr)$. So $\gamma_j^i(d_\star)$ and $\gamma_q^p(d_\star)$ are oriented inconsistently, $\vec{n}_i\big[s_i(\bldr)\big] = - \vec{n}_p\big[s_p(\bldr)\big]$, $\rho_2(\bldr) = \rho_1(\bldr) - 2 d_\star \vec{n}_i[s_i(\bldr)]\; \forall \bldr \in \mathfrak{R}$, and the segments $\gamma_j^i(2 d_\star)$, $\gamma_q^p$ have infinitely many common points. By Lem.~\ref{lem.anal}, $s_i \in \overset{\circ}{\gamma}\vphantom{\gamma}_j^i \mapsto \varrho_i(s_i|2d_\star)$ is a partial parametrization of $\gamma_q^p$. Given $(i,j)$ and $(p,q)$, this ``parametrization property'' holds for no more than one value of $d_\star$. So the case 3 holds for at most finitely many $d_\star$'s. This completes the proof.
\end{IEEEproof}
\begin{lemma}
There are no more than finitely many values of $d_\star$ for which iii) in Defn.~\ref{def.pliral} holds.
\label{i.fm2}
\end{lemma}
\begin{IEEEproof}
By Lem.~\ref{i.fm} and \ref{i.fm1}, we may focus on a value of $d_\star$ for which i) and ii) in Defn.~\ref{def.pliral} do not hold. By the concluding part of the proof of  Lem.~\ref{i.fm1}, we can also exclude $d_\star$ for which there exist $(i,j), (p,q)$ such that $s_i \in \overset{\circ}{\gamma}\vphantom{\gamma}_j^i \mapsto \varrho_i(s_i|2d_\star)$ is a partial parametrization of an analytic curve $C_q^p$ hosting $\gamma_q^p$. It suffices to show that there are no more than finitely many bridges with a length of $2d_\star$.
\par
Suppose the contrary. Then there exists an infinite sequence $\{[\bldp_1(k), \bldp_2(k)]\}_{k=1}^\infty$ of pairwise distinct bridges with $\|\bldp_1(k)-\bldp_2(k)\| = 2 d_\star\; \forall k$.
Properly thinning out this sequence results in a situation where for any $\nu=1,2$, either $\bldp_1(k)$'s are common for all $k$ or they are pair-wise distinct and lie in the interior $\overset{\circ}{\gamma}\vphantom{\gamma}_j^i$ of a common segment $\gamma_j^i$. Now we consider separately two feasible cases.
\par
{\bf Case 1:} For some $\nu=1,2$ (say $\nu=1$) all $\bldp_\nu(k)$'s are common $\bldp_1(k) = \bldp \; \forall k$, whereas $\bldp_2(k)$'s are pairwise distinct.
All they lie on the $2d_\star$ circle centered at $\bldp$. As before, we infer that $\gamma_j^i$ is a circular arc. It remains to note that the number of such circular segments  does exceed the number of all segments and so is finite.
\par
{\bf Case 2:} For $\nu=1,2$ all $\bldp_\nu(k)$'s are pairwise distinct. There exist $(i,j)$ and $(p,q)$ such that $\bldp_1(k) \in \overset{\circ}{\gamma}\vphantom{\gamma}^i_j, \bldp_2(k) \in \overset{\circ}{\gamma}\vphantom{\gamma}^p_q \; \forall k$. As before, we infer that  $s_i \in \overset{\circ}{\gamma}\vphantom{\gamma}_j^i \mapsto \varrho_i(s_i|2d_\star)$ is a partial parametrization of a curve $C_q^p$ hosting $\gamma_q^p$. However, this case was excluded.
\end{IEEEproof}
\par
{\it Proof of Proposition}~\ref{lem.bizarre}. This proposition is immediate from Lem.~\ref{i.fm}, \ref{i.fm1}, and \ref{i.fm2}.
\IEEEQED

\section{Proofs of Propositions~\ref{lem.fence} and \ref{lem.pere}.}
\label{app.fence}
We adopt their assumptions and, for $d_\star:=d_0,d_0+R, d_0+\rop$, use $d_\star$-ensembles from Defn.~\ref{ensemble}. By further partitioning the segments (if necessary), we ensure that these ensembles have a common base $\gamma_j^i$. By \eqref{prereq} and Defn.~\ref{def.pliral}, the second option from {\bf s.1)} in Defn.~\ref{ensemble} does not hold. Any segment $\gamma_j^i(d_\star)$ born by a point is said to be {\it circinate}, and {\it non-circinate} otherwise.
By partitioning the elements of the ensemble, if necessary, we ensure that the outer perimeter $\mathscr{P}$ is the cyclic concatenation of segments $\mathpzc{p}_1, \ldots, \mathpzc{p}_K$ such that
\begin{enumerate}[{\bf p1)}]
\item any $\mathpzc{p}_\nu$ equals $\gamma_j^i(d_0)$ for some $(i,j)=[i(\nu),j(\nu)]$;
\item for any extended object $D_i$, the following holds:
\begin{enumerate}
\item $1+d_0 \varkappa_i(s_i) > 0$ on $\overset{\circ}{\gamma}\vphantom{\gamma}^i_j$ for $(i,j):=[i(\nu),j(\nu)]$,
\item $\rext$ is to the right of $\varrho_i(s_i|d_0)$ as $s_i$ ascends,
\item $\pr{\varrho_i(s_i|d_0)}{D} = \{ \varrho_i(s_i) \}$ for the points on $\overset{\circ}{\gamma}\vphantom{\gamma}^i_j$;
\end{enumerate}
\item if a circinate segment $\mathpzc{p}_\nu$ is born by $\bldp$,\! then\!
$\pr{\overset{\circ}{\mathpzc{p}}_\nu}{D} \!\!= \!\!\{ \bldp \}$;
\item except for the points of concatenation in the cyclic order, different segments $\mathpzc{p}_\nu$'s have no common points.
\end{enumerate}
\par
The {\it normal leg} of $\bldp \in \mathpzc{p}_\nu$ is the segment $[\varrho_i(s_i), \bldp]$ (where $s_i$ is the root of $\varrho_i(s_i|d_0) = \bldp$) if $\mathpzc{p}_\nu$ is a non-circinate segment $\gamma_j^i(d_0)$, and the segment $[\bldp_i, \bldp]$ if $\mathpzc{p}_\nu$ is circinate and born by the point $\bldp_i$. The end of the normal leg different from $\bldp$ is called the {\it foot} of this leg.
\begin{lemma}
\label{lem.uniq}
Let $\bldp \in \mathpzc{p}_\nu$ and $[\rho,\bldp]$ be a normal leg of $\bldp$. Then
the following statements hold:
\begin{enumerate}[{i)}]
\item $\rho \in \partial D$, and $\|\bldr-\rho\|=d_0$;
\item $\dist{\bldr}{D} = \|\bldr-\rho\| \; \forall \bldr \in [\bldp,\rho]$;
\item  If the point $\bldp$ is not an end of the segment $\mathpzc{p}_\nu$, then the projection $\pr{\bldr}{D}=\{\rho\}$ for any $\bldr \in [\bldp,\rho]$.
\end{enumerate}
\end{lemma}
\begin{IEEEproof}
{\bf i)} is evident.
\par
{\bf ii)} We first note that
\begin{multline*}
\bldp \in \mathpzc{p}_\nu \subset \mathscr{P} \subset \partial \mathscr{N}^{d_0}(D)
\\
\Rightarrow \dist{\bldp}{D}=d_0 = \|\bldp - \rho\| \Rightarrow \rho \in \pr{\bldp}{D}.
\end{multline*}
So ii) is immediate from i) in Lem.~\ref{lem.distt}.
\par
{\bf iii)} Let $\bldr \in [\bldp,\rho]$ and $\rho_1, \rho_2 \in \pr{\bldr}{D}$. Then $\rho_i \in D,
\|\rho_i - \bldr \| = \dist{\bldr}{D} = \|\bldr - \rho\|$ and so
\begin{multline*}
d_0\leq \|\bldp - \rho_i\| \leq \|\bldp - \bldr\|+\|\bldr - \rho_i\|
\\
 = \|\bldp - \bldr\|+\|\bldr - \rho\| = \|\bldp - \rho\| = d_0.
\end{multline*}
Hence $\|\bldp - \rho_i\| = \dist{\bldp}{D} \Rightarrow \rho_i \in \pr{\bldp}{D}\; i=1,2$. The proof is completed by {\bf p2.c)} and {\bf p3)}.
\end{IEEEproof}
\begin{lemma} The following statements hold:
\begin{enumerate}[{\bf i)}]
\item The signed curvature $\kappa$ of any segment $\mathpzc{p}_\nu$ does not exceed $d_0^{-1}$;
\item any segment $\mathpzc{p}_\nu$ contains no more than finitely many roots of the equation $\kappa = - 1/R$;
\item If two such segments have a common end-point $\bldp$, either {\bf a)} their directed tangents at this point are identical or {\bf b)} the ``following'' tangent results from turning the ``preceding'' tangent to the right.
\end{enumerate}
\label{lem.fgfg}
\end{lemma}
\begin{IEEEproof}
{\bf i)} The curvature of the circinate segments equals $d_0^{-1}$.
For any other segment, Lem.~\ref{lem.curv} states that on $\overset{\circ}{\mathpzc{p}}_\nu$, the curvature $\kappa=\frac{\varkappa_i(s_i)}{1+d_0 \varkappa_i(s_i)}$, where  $1+d_0 \varkappa_i(s_i) > 0$ by {\bf p2.a)}. Hence $\kappa < d_0^{-1}$. By continuity, $\kappa \leq d_0^{-1}$ everywhere on $\mathpzc{p}_\nu$.
\par
{\bf ii)}
Any circinate segment has a positive curvature and so hosts no roots. For any other segment, any root $\mathpzc{r}$ is given by $\mathpzc{r} = \varrho_i(s_i|d_0)$, where
$$
\frac{\varkappa_i(s_i)}{1+d_0 \varkappa_i(s_i)} = \kappa = - R^{-1} \Rightarrow \varkappa_i(s_i) = - (d_0+R)^{-1}.
$$
The proof is completed by \eqref{prereq} and i) in Defn.~\ref{def.pliral}
\par
{\bf iii)} Suppose that this claim is untrue. Then the ``following'' tangent results from turning the ``preceding'' tangent to the left. We consider two points $\bldp^+ \neq \bldp^-$, each on its own segment from the concerned two ones. If they are close enough to $\bldp$, their normal legs intersect at a point $\bldr$ close to $\bldp$.
By Lem.~\ref{lem.uniq}, $\pr{\bldr}{D}$ is a singleton that contains the foot of any of these legs. So they are equal, then the legs are equal as well. Hence $\bldp^+ =\bldp^-$, in violation of the above property. This contradiction completes the proof.
\end{IEEEproof}
\par
Now we invoke the notations \eqref{def.rp}.
\begin{lemma}
Suppose that $\bldr \in \rext$ and $\rho \in \pr{\bldr}{\mathscr{P}}$.
\begin{enumerate}[{\bf i)}]
\item The outer perimeter $\mathscr{P}$ is a regular curve at the point $\rho$ and the vector $\bldr - \rho$ is normal to $\mathscr{P}$ at $\rho$.
\item $\pr{\rho}{D}$ contains a single point $\bldp$ and
\begin{equation}
\label{one-to-one}
\rho = \bldr_{\bldp \vartriangleright d_0}, \quad \bldr = \bldr_{\bldp \vartriangleright (d_0 +\|\bldr-\rho\|)}.
\end{equation}
\item $\bldr - \rho$ and $\rho - \bldp$ are co-linear and identically oriented.
\end{enumerate}
\label{lem.twoproj}
\end{lemma}
\begin{IEEEproof}
{\bf i)} follows from items ii,iii) in Lem.~\ref{for.norra} and item iii) in Lem.~\ref{lem.fgfg}.
\par
{\bf iii)} Let $\bldp \in \pr{\rho}{D}$. Then $\|\rho - \bldp\| = d_0$, whereas $\|\bldr - \rho\| = \dist{\bldr}{\mathscr{P}}$. If iii) fails to be true,
\begin{multline*}
\dist{\bldr}{D} \leq \|\bldr - \bldp\| <
\|\bldr - \rho\| + \|\rho - \bldp\| \\ = \dist{\bldr}{\mathscr{P}}+d_0,
\end{multline*}
in violation of Prop.~\ref{lem.chch0}. This contradiction completes the proof of iii).
\par
{\bf ii)} is immediate from iii).
\end{IEEEproof}
\begin{cor}
\label{cor.size} For $\bldr \in \rext$, there is a one-to-one correspondence between the projections $\pr{\bldr}{\mathscr{P}}$ and $\pr{\bldr}{D}$; this correspondence is given by the first formula in \eqref{one-to-one}.
\end{cor}
\par
An $R$-disc $\mathb$ is said to be {\it touching} if its interior
$\mathbf{int}\mathb$ lies in $\rext$ and the {\it contact patch} $T(\mathb):=\mathb \cap \mathscr{P}$ is not empty. Its {\it main angle} $A(\mathb)$ is rooted at the center of $\mathb$, covers $T(\mathb)$ and, among all such angles, has the minimal angular measure. If the contact patch $T(\mathb)$ is a singleton, the angle $A(\mathb)$ has zero measure, i.e., degenerates into a ray.
A touching $R$-disc $\mathb$ is said to {\it roll} on $\mathscr{P}$ if it continuously depends on a real parameter and the contact patch $T(\mathb)$ always contains a single (contact) point $\bldp$, which may be different for every parameter.
The $\ve$-{\it displaced center} of a rolling disc is given by $(R+\ve)N(\bldc_{\mathb}, \bldp)$, where $\bldc_{\mathb}$ is the center of $\mathb$ and $N(\cdot,\cdot)$ is defined in \eqref{def.rp}.
\par
\begin{figure}
\centering
\scalebox{0.2}{\includegraphics{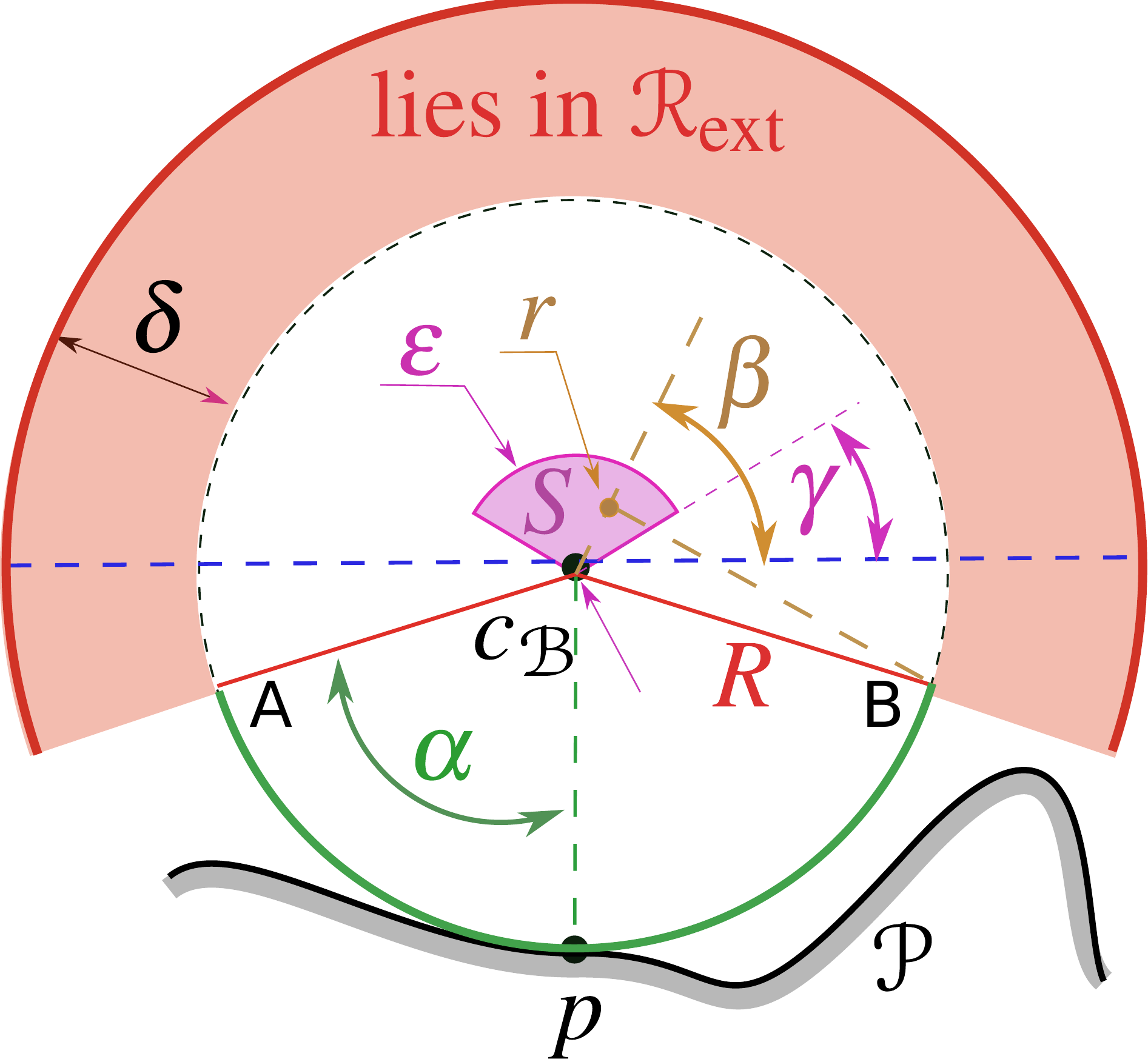}}
\caption{ A sector covered by $Q(R)$.}
\label{fig.dc}
\end{figure}

The next lemma uses the notation \eqref{def.q}.
\begin{lemma}
\label{lemtouch}
For any touching $R$-disc $\mathb$, the following statements are true:
\begin{enumerate}[{\bf i)}]
\item \label{lemtouch.i} The contact patch $T(\mathb)$ is a finite set;
\item \label{lemtouch.ii} The vector $\bldc_{\mathb}-\bldp$ is normal to the outer perimeter $\mathscr{P}$ at any point $\bldp \in T(\mathb)$;
\item \label{lemtouch.iii} If $T(\mathb)$ contains a single point $\bldp$, then
\begin{enumerate}
\item the disc $\mathb$ can roll on $\mathscr{P}$ in any direction until the size of the contact patch $T(\mathb)$ jumps up;
\item For any $\gamma \in (0,\pi/2)$, there is $\ve>0$ such that during an initial period of motion, the (pink) sector $S$ from Fig.~\ref{fig.dc} lies in the set $Q(R)$ given by \eqref{def.q}.
\end{enumerate}
\end{enumerate}
\end{lemma}
\begin{IEEEproof}
{\bf i)} Suppose the contrary. Then there is a segment $\mathpzc{p}_\nu$ with infinitely many points in common with the boundary circle $\partial \mathb$. By Lem.~\ref{lem.anal}, $\mathpzc{p}_\nu$ is a part of $\partial \mathb$ and so its curvature is negative. Since the ``point-born'' segments $\gamma_j^i(d_0)$ have positive curvatures, $\mathpzc{p}_\nu = \gamma_j^i(d_0)$ for an extended object $D_i$ and $j$ such that {\bf p2.a)} holds: $1+d_0 \varkappa_i(s_i) > 0$ on $\overset{\circ}{\gamma}\vphantom{\gamma}^i_j$. Then by Lem.~\ref{lem.curv},
$$
R^{-1} = \frac{-\varkappa_i(s_i)}{1+d_0 \varkappa_i(s_i)} \Rightarrow \varkappa_i(s_i) = - \frac{1}{d_0+R}\; \forall s_i \in \overset{\circ}{\gamma}\vphantom{\gamma}^i_j.
$$
 By i) in Defn.~\ref{def.pliral}, this means that the distance $d_0+R$ is bizarre, in violation of \eqref{prereq}. This contradiction proves i).
\par
{\bf ii)} The claim is straightforward from i) in Lem.~\ref{lem.twoproj}.
\par
{\bf iii.a)} Let $\mathpzc{p}_\nu$ be the segment that contains a subsegment $\mathpzc{p} \ni \bldp$ in the desired direction of motion on $\mathscr{P}$, and let $\bldc$ be the center of $\mathb$. The $R$-circle $\partial \mathb$ is tangent to $\mathpzc{p}_\nu$ at $\bldp$ by iii) in Lem.~\ref{lem.fgfg}, and $\mathbf{int}\mathb \cap \mathpzc{p}_\nu = \emptyset$.
Hence $\dist{c}{\mathscr{P}} = R$ and by ii-iv) in Lem.~\ref{for.norra} (and by reducing $\mathpzc{p}$, if necessary), the $R$-disc $\mathb(\widetilde{\bldp})$ centered at $\bldc(\widetilde{\bldp}):= \widetilde{\bldp} - R \vec{n}_{\mathscr{P}}(\widetilde{\bldp})$ has only one point in common with $\mathpzc{p}$ whenever $\widetilde{\bldp} \in \mathpzc{p}, \widetilde{\bldp} \neq \bldp$.
\par
By reducing $\mathpzc{p}$ also further, it can be ensured that for any $\widetilde{\bldp} \in \mathpzc{p}$, the perimeter $\mathscr{P}$ and $\mathb(\widetilde{\bldp})$ have a single common point $\widetilde{\bldp}$. Indeed, otherwise, there exists an infinite sequence $\{ \bldp_k \}_{k} \in \mathpzc{p}$ such that $\bldp_k \to \bldp$ as $k \to \infty$ and the disc $\mathb(\bldp_k)$ contains a point $\bldr_k \in \mathscr{P}, \bldr_k \neq \bldp_k$. By the foregoing, $\bldr_k \not\in \mathpzc{p}$. By passing to a subsequence, we ensure that $\exists \bldr := \lim_{k \to \infty} \bldr_k$. Then $\bldr \in \mathscr{P} \cap \mathb$ and $\bldr \not\in \mathpzc{p} \ni \bldp \Rightarrow \bldr \neq \bldp$, in violation of $\mathscr{P} \cap \mathb = \{\bldp\}$. Hence $\mathscr{P} \cap \mathb(\widetilde{\bldp}) = \{ \widetilde{\bldp}\}$.
\par
It follows that the disc can roll on $\mathscr{P}$ until the size of the contact patch $T(\mathb)$ jumps up.
\par
{\bf iii.b)} We pick $\alpha \in (0,\gamma)$ and note that for a sufficiently small $\delta>0$,
the set $\rext$ contains the red sector of the annulus from Fig.~\ref{fig.dc}. Let $\ve \in (0,\delta)$.
The distance from $\bldr \in S$ to $\mathscr{P}$ is no less than $R+\delta - \ve >R$ along any ray that goes from $\bldr$ through the red arc in Fig.~\ref{fig.dc}.
Elementary planimetry shows that along any other ray, that distance is no less than the distance $d$ to the set $ W= \{ A , B \}$.
By looking at the triangle whose vertices are $\bldr, \bldc_{\mathb}$ and $\pr{\bldr}{W}$, we see that
\begin{gather*}
d = \sqrt{R^2 + \|\bldr - \bldc_{\mathb}\|^2 - 2 R \|\bldr - \bldc_{\mathb}\| \cos (\pi/2 - \alpha + \beta)}
\\
 = \sqrt{R^2 + \|\bldr - \bldc_{\mathb}\|^2 + 2 R \|\bldr - \bldc_{\mathb}\| \sin (\beta - \alpha)}
 \\
 \geq \sqrt{R^2 + \|\bldr - \bldc_{\mathb}\|^2 + 2 R \|\bldr - \bldc_{\mathb}\| \sin (\gamma - \alpha)} >R.
\end{gather*}
It remains to note that $\delta$ and $\ve \in (0,\delta)$ can be chosen common for an initial period of motion.
\end{IEEEproof}
\par
\begin{figure}
\begin{center}
\subfigure[]{\scalebox{0.2}{\includegraphics{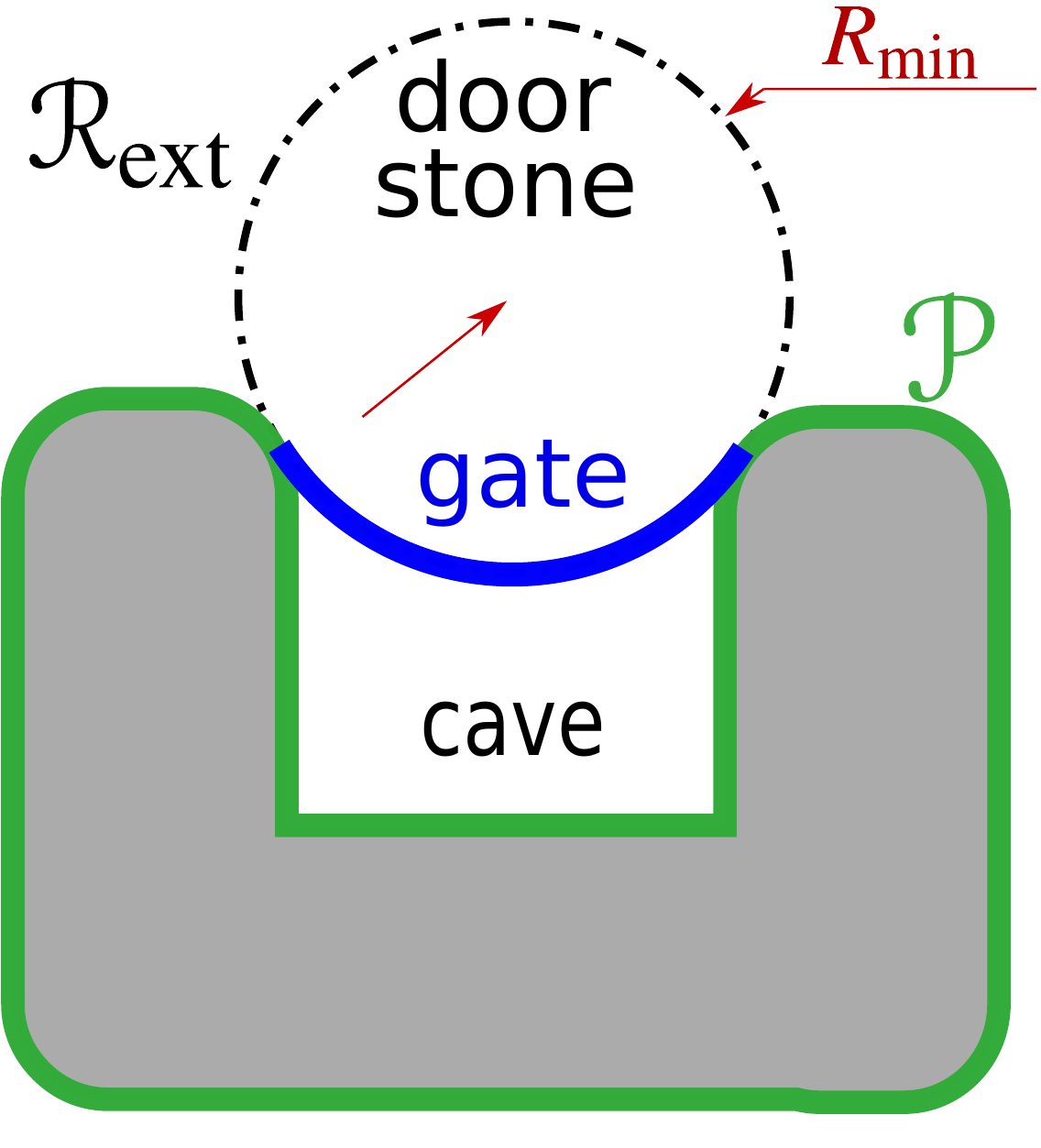}}\label{fig.door1}}
\hfil
\subfigure[]{\scalebox{0.2}{\includegraphics{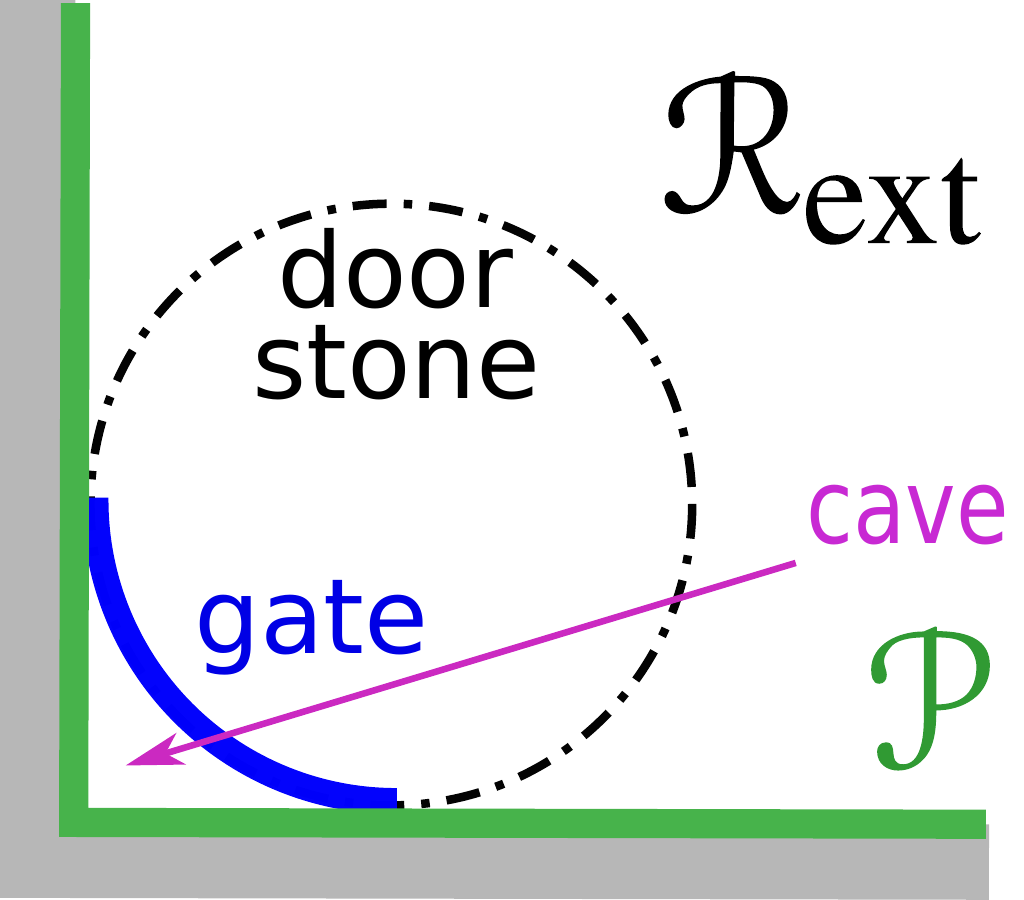}}\label{fig.door2}}
\hfil
\subfigure[]{\scalebox{0.2}{\includegraphics{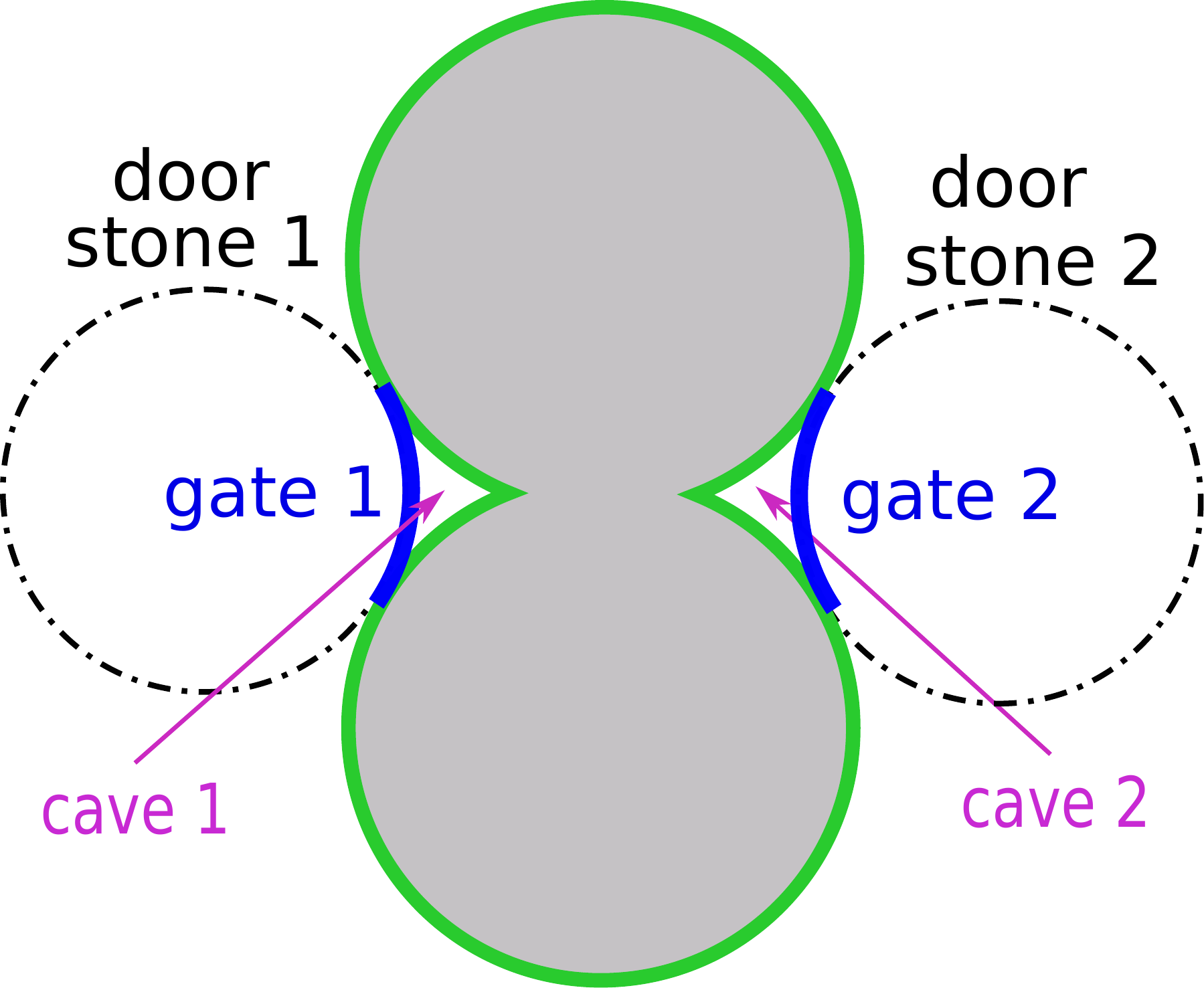}}\label{fig.door3}}
\end{center}
\caption{Door stones and gates.}
\label{fig.door}
\end{figure}
\begin{figure}
\centering
\scalebox{0.2}{\includegraphics{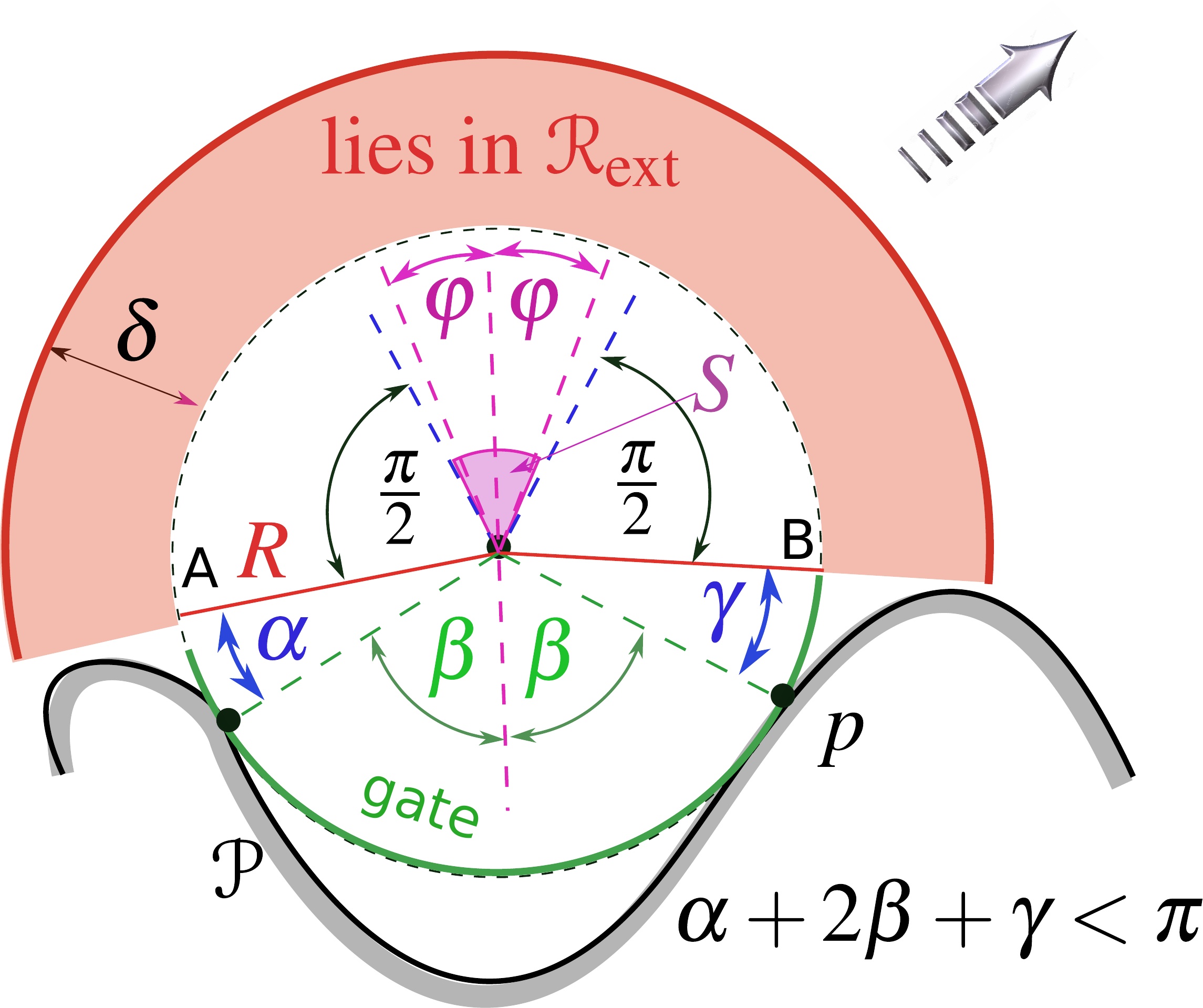}}
\caption{A sector covered by $Q(R)$.}
\label{fig.dc1}
\end{figure}

By \ref{lemtouch.iii}) in Lem.~\ref{lemtouch}, the rolling motion of $\mathb$ on $\mathscr{P}$ either is endless or ends in a situation where $\mathb$
hits $\mathscr{P}$ at many points. They are hosted by a (minimal) arc $\mathscr{A}_{\mathb}$ of $\partial \mathb$ with an angular measure $\leq \pi$. Moreover, this measure is less than $\pi$ since otherwise, the base chord, extended up to hitting $D$, is a $(d_0+R)$-bridge due to iii) in Defn.~\ref{def.pliral}, i) in Lem.~\ref{for.norra} and \ref{lemtouch.ii}) in Lem.~\ref{lemtouch}, in violation of \eqref{prereq}.
Any arc of $\mathscr{A}_{\mathb}$ between two consecutive contact points is called the {\it gate} (to a cave); the {\it cave} is the area bounded by the gate and the part of $\mathscr{P}$ that is run between these points so that $\rext$ is to the right. The disc $\mathb$ is then called the {\it door stone}; see. Fig.~\ref{fig.door}.
There are finitely many door stones and points, where they touch $\mathscr{P}$. By partitioning $\mathpzc{p}_\nu$'s, we ensure that no such point belongs to $\overset{\circ}{\mathpzc{p}}_\nu$.
\begin{lemma}
\label{lem.fromcave}
Let $\mathb$ be a door stone and let
$\bldp$ be an end of the arc $\mathscr{A}_{\mathb}$.
\begin{enumerate}[{\bf i)}]
\item
The touching disc $\mathb$ can roll on the outer perimeter $\mathscr{P}$, at least a bit, so that the touching point starts from $\bldp$ and goes away from the $\mathb$-caves;
\item There exist $\varphi >0 $ and $\ve>0$ such that during an initial period of motion, the (pink) sector $S$ of the $\ve$-disc from Fig.~\ref{fig.dc1} lies in the set $Q(R)$ given by \eqref{def.q}.
\end{enumerate}
\end{lemma}
\begin{IEEEproof}
{\bf i)} We pick a segment $\mathpzc{p}_\nu \ni \bldp$ that goes outside the $\mathb$-caves. By \ref{lemtouch.ii}) in Lem.~\ref{lemtouch}, $\mathscr{P}$ and $\partial \mathb$ are tangent at $\bldp$. So by ii-iv) in Lem.~\ref{for.norra}, the disc can roll on a subsegment $\mathpzc{p} \subset \mathpzc{p}_\nu$ so that the touching point goes away from the caves. We intend to show that in fact, the disc rolls on $\mathscr{P}$, i.e., has a single point in common with $\mathscr{P}$.
\par
Let the disc roll without slipping and let its center move with the unit speed; the symbol $\mathb(t)$ stands for its position at time $t$.
At the start, the instantaneous center of rotation is $\bldp$ and the angular speed equals $R^{-1}$ in absolute value. Hence the velocity vectors of all points from $\mathscr{A}_{\mathb} \setminus \bldp$ are initially directed inside $\mathb$. The straight segments that link $\bldc_{\mathb}$
and the ends of $\mathscr{A}_{\mathb}$ are not co-linear.
It follows that $\mathscr{A}_{\mathb}$ can be extended beyond the end different from $\bldp$ so that the above velocity direction holds on the extended arc $\mathscr{A}_\ast$ (except for $\bldp$). Hence
the position $\mathscr{A}_\ast(t)$ of $\mathscr{A}_\ast$ is inside $\mathb(0)$ for $t \approx 0, t>0$.
\par
Suppose that the disc does not roll on $\mathscr{P}$, even a bit. Then there are infinitely many points $\{ \bldr_k\}$ and times $\{t_k\}$ such that $\bldr_k \not\in \mathpzc{p}, \; \bldr_k\in \mathscr{P} \cap \mathb(t_k) \; \forall k$ and $t_k \to 0+$ as $k \to \infty$. By passing to a subsequence, we ensure that $\exists \bldr_\infty := \lim_{k \to \infty} \bldr_k$. Then the foregoing guarantees that $\bldr_\infty \in \mathscr{P}$, $\bldr_\infty \not\in \mathpzc{p} \cup  \mathscr{A}_\ast$, whereas $\mathscr{P} \cap \mathb \subset \mathpzc{p} \cup \mathscr{A}$. This contradiction completes the proof of i).
\par
{\bf ii)} In Fig.~\ref{fig.dc1}, $\beta \in (0,\pi/2)$.
We pick $\alpha, \gamma >0$ so that $\alpha+2 \beta+\gamma < \pi$
For a small $\delta>0$, the set $\rext$ contains the red sector of an annulus from Fig.~\ref{fig.dc1}.
Let $\ve \in (0,\delta)$ and $\varphi < \frac{1}{2} [\pi - \alpha - 2 \beta-\gamma ]$.
The proof is completed like in the case ii.b) of Lem.~\ref{lemtouch}.
\end{IEEEproof}
\par
If a rolling disc $\mathb$ gets into a situation with many contact points, it can continue rolling due to
Lem.~\ref{lem.fromcave}. In this situation, $\mathb$ is a door stone, and the boundaries of the $\mathb$-caves are composed by arcs of $\partial \mathb$ and consecutive segments from the family of $\mathpzc{p}_\nu$'s. Passing through this situation can be viewed as if the disc and the contact point instantaneously run over the concerned segments and arcs, respectively.
Thus, the disc eventually returns to the initial position, with having swept through the entire sequence of the segments $\mathpzc{p}_1, \ldots, \mathpzc{p}_K$ forming $\mathscr{P}$.
The {\it central} $\mathscr{C}$ and {\it contact} $\mathscr{T}$ paths are those run by the disc center and the contact point, respectively, during this motion.
\begin{lemma}
\label{lem.mm}
The following statements hold:
\begin{enumerate}[{\bf i)}]
\item Central paths are the boundaries of the connected components $Q_{cc}$ of the set \eqref{def.q};
\item These paths are piece-wise analytic Jordan loops that meet {\bf p1--p4)} with $\rext \mapsto Q_{\text{cc}}$ in {\bf p2.b)} and $d_0\! \mapsto \!d_0+R$;
\item Any contact path $\mathscr{T}$ is a major $(R,\rop)$-loop;
\item Any contact path $\mathscr{T}$ bounds the union of the $R$-discs centered at the component $Q_{cc}$ encircled by the related central path $\mathscr{C}$, and $\dist{\bldp}{\mathscr{C}} = R \; \forall \bldp \in \mathscr{T}$;
\item If $\bldr$ lies in a $R$-disc $\mathpzc{D} \subset \rext$, then $\bldr$ lies in the patch of a major $(R,\rop)$-loop.
\end{enumerate}
\end{lemma}
\begin{IEEEproof}
{\bf i)} Let $\sa{\vec{w}}{\vec{g}}$ stand for the angle between the nonzero vectors $\vec{w}$ and $\vec{g}$. Any central path $\mathscr{C}$ contains no more than finitely many consecutive positions $\bldc_1^0, \ldots, \bldc_N^0$ of the center of the rolling disc for which this disc is a door stone. Thanks to ii) in Lem.~\ref{lem.fromcave}, there exists $\ve>0$ such that for any $i=1,\ldots,N$, the set $Q(R)$ covers a pink sector with an angular measure of $2 \varphi_i$ around it, like in Fig.~\ref{fig.dc1}. So it also covers  the straight segment $(\bldc_i^0, \bldc_i^0+ \ve \vec{w}_i]$, where $ \vec{w}_i=1$ is the unit centerline vector of this sector.
We put $\varphi:= \min_i \varphi_i$, pick $\gamma >0$ so small that
$$
\sa{\vec{w}_i}{\bldc_i - \bldp} < \pi/2 - \gamma \; \forall \bldp \in \pr{\bldc_i}{\mathscr{P}}, \forall i
$$
and
build a continuous vector-field $\bldc \in \mathscr{C} \mapsto \vec{w}(\bldc)$ such that $\|\vec{w}(\bldc)\|=1, \vec{w}(\bldc_i^0) = \vec{w}_i \; \forall i$, and $\sa{\vec{w}(\bldc)}{\bldc - \bldp(\bldc)} < \pi/2 - \gamma$ whenever $\bldc \neq \bldc_i^0 \; \forall i$, where $\bldp(\bldc) \in \pr{\bldc}{\mathscr{P}}$. We are going to show that
\begin{equation}
q_\delta(\bldc):= \bldc + \delta \vec{w}(\bldc) \in Q(R)\quad \forall \bldc \in \mathscr{C}, \delta \approx 0.
\label{appr.path}
\end{equation}
\par
Suppose the contrary. Then  there exists two sequences $\{\delta_k\}_{k} \subset (0,\infty)$ and $\{\bldc_k\}_{k} \subset \mathscr{C}$ such that $\delta_k \to 0$ as $k \to \infty$ and $q_\delta(\bldc_k) \not\in Q(R) \; \forall k$. By passing to a subsequence, we ensure that there exists $\bldc := \lim_{k \to \infty} \bldc_k$. Then iii.b) in Lem.~\ref{lemtouch} and ii) in Lem.~\ref{lem.fromcave} imply that
\begin{gather*}
\sa{\vec{w}(\bldc_k)}{\bldc_k - \bldp_k} \geq \pi/2 - \gamma\; \forall k \approx \infty \quad \text{if}\quad \bldc \neq \bldc_i^0 \; \forall i;
\\
\sa{\vec{w}(\bldc_k)}{\bldc_k - \bldp_k} \geq \varphi\; \forall k \approx \infty \quad \text{if} \quad \bldc = \bldc_i^0 \quad \text{for some} \; i,
\end{gather*}
where $\bldp_k \in \pr{\bldc_k}{\mathscr{P}}$. Letting $k \to \infty$ here yields a contradiction with the defining property of the vector-field $\vec{w}(\cdot)$ at the point $\bldc$.
\par
Due to \eqref{appr.path}, $q_\delta(\bldc)$ lies in a certain connected component $Q_{cc}$ of $Q(R)$ for all $\bldc \in \mathscr{C}$ and $\delta \approx 0$. Letting $\delta \to 0+$ shows that $\mathscr{C} \subset \ov{Q}_{cc}$, whereas $\dist{\bldc}{\mathscr{P}} = R \; \forall \bldc \in \mathscr{C} \Rightarrow \mathscr{C} \cap Q(R) = \emptyset$. Thus $\mathscr{C} \subset \partial Q_{cc}$ and so the curve $\mathscr{C}$ is closed and does not intersect itself.
\par
Suppose that $\mathscr{C} \neq \partial Q_{cc}$.
Then the connected set $Q_{\text{cc}}$ contains a ``hole'' $H$. We pick $\bldc \in H$ and $\bldp \in \pr{\bldc}{\mathscr{P}}$. For $\bldp^\prime \in [\bldp, \bldc]$, we have
$\dist{\bldp^\prime}{\mathscr{P}} = \|\bldp^\prime - \bldp\| \leq \|\bldc - \bldp\| < R$ by i) in Lem.~\ref{lem.distt}. However, as the point $\bldp^\prime\in [\bldp, \bldc]$ continuously moves from $\bldc$ to $\bldp$, it must leave the hole and thus arrive at a position $\bldp^\prime \in \partial Q_{\text{cc}}$, where $ \|\bldc - \bldp\| = R$. This contradiction proves that $\mathscr{C} = \partial Q_{cc}$.
\par
Conversely, let $Q_{\text{cc}}$ be a connected component of the set \eqref{def.q}. We pick $\bldr \in Q_{cc}$ and $\bldp \in \pr{\bldr}{\mathscr{P}}$.
By i) in Lem.~\ref{lem.distt}, there is $\bldc \in [\bldp,\bldr]$ such that $\|\bldc - \bldp\| =R$. Then $\bldc \in \partial Q_{\text{cc}}$ and the $R$-disc centered at $\bldc$ is touching. By rolling it over $\mathscr{P}$, we get a central path $\mathscr{C}$ such that $\mathscr{C} \cap \partial Q_{\text{cc}} \neq \emptyset$ and so $\mathscr{C} \subset \partial Q_{\text{cc}}$. Hence $\mathscr{C} = \partial Q_{\text{cc}}$, as before.
\par
{\bf ii)}
Since not only $d_0$ but also $d_0+R$ is not bizarre, the arguments on the structure of $\mathscr{P}$ from the beginning of App.~\ref{app.fence} extend on $\partial Q_{\text{cc}}$.
\par
{\bf iii)} Any contact path $\mathscr{T}$ is a loop built of cyclically concatenated segments that are either 1) from the family of $\mathpzc{p}_\nu$'s or 2) are arcs of the boundary of a door stone.
In this loop, any two consecutive segments of type 1 are tangential at the common end by iii) in Lem.~\ref{lem.fgfg}. If segments of types 1 and 2 are in contact, then the boundary of the parent door stone $\mathb$ of the latter is tangent to the former at the common end by the same argument, and so the segments are tangential.
Any consecutive segments of type 2 are tangential since they lie on a common circle $\partial \mathb$. Thus we see that $\mathscr{T}$ is a regular curve.
\par
In absolute value, the curvature of any segment of type 2 equals $R^{-1} \leq \rop^{-1}$. For the segments of type 1, the signed curvature does not exceed $d_0^{-1} \leq \rop^{-1}$ by i) in Lem.~\ref{lem.fgfg} and \eqref{hidden.ass}, whereas it is no less than $-R^{-1}$ by ii,iii) in Lem.~\ref{for.norra}. So $\mathscr{T}$ is a $(R,\rop)$-loop by Defn.~\ref{rmin.loop2}.
\par
{\bf iv)}
The central $\mathscr{C}$ and contact $\mathscr{T}$ paths are born by a common motion of a $R$-disc $\mathb$. Any point $\bldp \in \mathscr{T}$ lies at a distance of $\leq R$ from $\mathscr{C}$ and closer to $\mathscr{P}$. If $\dist{\bldp}{\mathscr{C}} < R$, then $\bldp$ lies in the interior of the disc $\mathb$ at some moment of its motion. However, this is infeasible since $\bldp$ lies on a segment of type 1 or 2. So $\dist{\bldp}{\mathscr{C}} = R$.
\par
By construction, $\mathscr{T}$ bounds the union of the $R$-discs centered at $\mathscr{C} = \partial Q_{cc}$. Let $\mathpzc{D}$ be a $R$-disc centered at $\bldc \in Q_{cc}$. Since $Q_{cc}$ is connected, the center $\bldc$ can be continuously moved inside $Q_{cc}$ to the boundary $\partial Q_{cc}$.
During the associated motion of the disc, its boundary touches $\mathscr{T}$ only at the very end. Hence except for this moment, the moving disc is always inside $\mathscr{T}$, including the initial moment. So $\mathscr{D}$ is inside $\mathscr{T}$, which completes the proof of iv)
\par
{\bf v)} It suffices to note that the center of $\mathscr{D}$ lies in $Q(R)$ and so in some of its connected components $Q_{cc}$, and to refer to iv).
\end{IEEEproof}
\par
{\it Proof of Proposition}~\ref{lem.fence}:
{\bf i)} is true due to i) and ii) in Lem.~\ref{lem.mm}.
\par
{\bf ii)} is true due to iii) and iv) in Lem.~\ref{lem.mm}.
\par
{\bf iii)} Let us consider a secure trajectory $\bldr(t)$ and the accompanying $\rop$-disc $\mathb(t)$. Then $\dist{\bldc(t)}{\mathscr{P}} > \rop$ for the center $\bldc(t)$ of $\mathb(t)$, and so $\bldc(t)$ evolves within a single connected component $Q_{\text{cc}}$.
The proof is completed by ii).
\par
{\bf iv)} The connected component $Q_{cc}(R)$ of $Q(R)$ that parents the major $(R,\rop)$-loop $\mathscr{F}$ at hands lies in $Q(R_\ast)$ and so in a connected component $Q_{cc}(R_\ast)$ of $Q(R_\ast)$. Any point $\bldr$ from the patch of $\mathscr{F}$ lies in an $R$-disc $\mathscr{D}(R) \subset \rext$ centered at $\bldc \in Q_{cc}(R) \subset Q_{cc}(R_\ast)$ by ii).
The $R_\ast$-disc $\mathscr{D}(R_\ast)$ centered at $\bldc$ can be continuously moved inside $\mathscr{D}(R)$ so that it eventually covers $\bldr$. The distance from its center to $\mathscr{P}$ constantly exceeds $R_\ast$ since $\mathscr{D}(R) \subset \rext$. Thus the center continuously moves inside $Q_{cc}(R_\ast)$ and so $\bldr$ lies in the patch of the major $(R_\ast,\rop)$-loop born by $Q_{cc}(R_\ast)$ by ii).
\par
If the patch of the major $(R,\rop)$-loop $\mathscr{F}$ at hands simultaneously lies in the patches born by two connected components $Q^\prime_{cc}(R_\ast)$ and $Q^{\prime\prime}_{cc}(R_\ast)$, then $Q_{cc}(R) \subset Q^\prime_{cc}(R_\ast), Q^{\prime\prime}_{cc}(R_\ast)$ and so $Q^\prime_{cc}(R_\ast) \cap Q^{\prime\prime}_{cc}(R_\ast) \neq \emptyset \Rightarrow Q^\prime_{cc}(R_\ast) = Q^{\prime\prime}_{cc}(R_\ast)$.
\IEEEQED
\par
{\it Proof of Proposition}~\ref{lem.pere}:
In Asm.~\ref{ass.init}, the center of the disc is at a distance $>R$ from $\mathscr{P}$. By ii) in Prop.~\ref{lem.fence}, this disc is covered by the patch of a major
$(R,\rop)$-loop; so are the circles $C_\pm^{\text{in}}$ by Asm.~\ref{ass.init}. It remains to note that $\bldr_{\text{in}}\in C_\pm^{\text{in}}$ and to invoke v) in Prop.~\ref{lem.fence}.

\bibliographystyle{IEEEtran}
  \bibliography{Hamidref}
 \end{document}